\pdfoutput=1
\documentclass[11pt,a4paper]{article}
\usepackage[latin1]{inputenc}
\usepackage{amsmath}
\usepackage{amsfonts}
\usepackage{amssymb}
\usepackage{authblk}
\usepackage[letterpaper, margin=1in]{geometry}
\usepackage{datetime}
\usepackage{array}
\usepackage{url}

\usepackage[numbers]{natbib}
\usepackage{graphicx}
\usepackage{subfig}
\usepackage{epstopdf}
\usepackage{multirow}
\usepackage{mathtools}
\usepackage{floatrow}
\usepackage{caption}
\usepackage{xr}
\makeatletter
\newcommand*{\addFileDependency}[1]{
  \typeout{(#1)}
  \@addtofilelist{#1}
  \IfFileExists{#1}{}{\typeout{No file #1.}}
}
\makeatother

\newcommand*{\myexternaldocument}[1]{
    \externaldocument{#1}
    \addFileDependency{#1.tex}
    \addFileDependency{#1.aux}
}

\myexternaldocument{supplement}

\newcommand{\lm}{\lambda}
\newcommand{\dt}{d_{\theta}}

\newcommand{\di}{d_{\phi}}

\providecommand{\keywords}[1]
{
  \small	
  \textbf{\textit{Keywords---}} #1
}

\author[1]{Abhinav Gupta\thanks{guptaa@mit.edu}}
\author[1]{Pierre F.J. Lermusiaux\thanks{pierrel@mit.edu (Corresponding author.)}}

\affil[1]{Department of Mechanical Engineering, Massachusetts Institute of Technology, Cambridge, MA 02139}

\date{\today}
\title{Neural Closure Models for Dynamical Systems}

\begin{document}

\date{\today}
\maketitle

\begin{abstract}
%
%
Complex dynamical systems are used for predictions in many domains.
%
Because of computational costs, models are truncated, coarsened, or aggregated. As the neglected and unresolved terms become important, the utility of model predictions diminishes.
%
We develop a novel, versatile, and rigorous methodology to learn non-Markovian closure parameterizations for known-physics/low-fidelity models using data from high-fidelity simulations. The new \textit{neural closure models} augment low-fidelity models with neural delay differential equations (nDDEs), motivated by the Mori-Zwanzig formulation and the inherent delays in complex dynamical systems. We demonstrate that neural closures efficiently account for truncated modes in reduced-order-models, capture the effects of subgrid-scale processes in coarse models, and augment the simplification of complex
biological
and physical-biogeochemical models.
%
We find that using non-Markovian over Markovian closures improves long-term prediction accuracy and requires smaller networks. 
We derive adjoint equations and network architectures needed to efficiently implement the new discrete and distributed nDDEs, for any time-integration schemes and allowing nonuniformly-spaced temporal training data.
%
The performance of discrete over distributed delays in closure models is explained using 
information theory, and we find an optimal amount of past information for a specified architecture. Finally, we analyze computational complexity and explain the limited additional cost due to neural closure models.
\end{abstract}


\section{Introduction}

Most models only resolve spatio-temporal scales, processes, and field variables to a certain level of accuracy because of the high computational costs associated with high-fidelity simulations.  Such truncation of scales, processes, or variables often limit the reliability and usefulness of simulations, especially for scientific, engineering, and societal applications where longer-term model predictions are needed to guide  decisions. There are many ways to truncate high-fidelity models to low-fidelity models. 
 Examples abound and three main classes of truncations are: evolving the original dynamical system in a reduced space, e.g., using  reduced-order-models (ROMs) \cite{wang2012proper,kutz_etal_DMD2016}); coarsening the model resolution to the scales of interest \cite{lesieur_etal_LES-2005,alfonsi2009reynolds}; and
reducing the complexity or number of state variables, components, and parameterizations  \cite{los2010complexity,chassignet_verron_ocean_param2012,ward2013biogeochemical}. 
In many applications, the neglected and unresolved terms along with their interactions with the resolved ones can become important over time, and a variety of modeling techniques have been developed to represent the missing terms. Techniques that express these missing terms as functions of modeled state variables and parameters are referred to as closure models. A main challenge is that no one closure approach to date is directly applicable to all four main classes of model truncations.  Another is that closure models are only well-defined for either linear problems or simple cases. Finally, they can easily become ineffective in the face of nonlinearities. 

Due to the explosion of use of a variety of machine learning methods for solving or simulating dynamical systems, a number of data-driven methods have been proposed for the closure problem. Most of them attempt to learn a neural network as the instantaneous map between the low-fidelity solution and the residual of the high- and low-fidelity solution, or their residual dynamics
\cite{pawar2020data, wan2018data, wang2020recurrent, san2018neural, pan2018data}. They often use recurrent networks such as
long-short term memory networks (LSTMs), gated recurrent units (GRUs) etc., with justification based on the Mori-Zwanzig formulation \cite{stinis2015renormalized,chorin2000optimal, gouasmi2017priori} and embedding theorems by Whitney \cite{whitney1936differentiable} and Takens \cite{takens1981detecting}.
These approaches do not however take into account accumulation of numerical time-stepping error in the presence of neural-networks during training.
and uniformly-spaced
high-fidelity data to be able to compute the time derivative of the state with high level of accuracy. 
Such requirement on the training data can be a luxury in a lot of scenarios. The requirement of very frequent snapshot data of the system is also true for methods which achieve model discovery using sparse-regression and provide interpretable learned models \cite{brunton2016discovering,kulkarni_et_al_DDDAS2020, pan2018data}.
All of the above issues are addressed by using neural ordinary differential equations (nODEs; \cite{chen2018neural}) and some researchers recently used nODEs for closure modeling. Some directly learn the ODE system from high-fidelity simulation data without using the available low-fidelity models \cite{maulik2020time}, which could lead to the requirement of bigger neural networks. Others combine nODEs with model discovery using sparse-regression \cite{yang2020bayesian} or only learn the values of parameters in existing closure models \cite{portwood2019turbulence}. 
Nearly all existing studies
primarily only attempt to address the closure for ROMs.
Finally, the existing machine learned closure models are not yet used for long-term predictions, i.e.\;forecasting significantly outside of the time-period to which the training data belonged to. 

In the present study, we propose a new neural delay differential equations (nDDEs) based framework to learn closure parameterizations for low-fidelity models using data from high-fidelity simulations and to increase the long-term predictive capabilities of these models. Instead of using ODEs, we learn non-Markovian closure models using DDEs. We base the theoretical justification for using DDEs on the Mori-Zwanzig formulation \cite{stinis2015renormalized, chorin2000optimal, gouasmi2017priori} and the presence of inherent delays in many dynamical systems \cite{otto2019nonlinear}, especially biological systems \cite{glass2020nonlinear, tokuda2019reducing}. We refer to the new modeling approach as \textit{neural closure models}. 
We demonstrate that our methodology drastically improves the predictive capability of low-fidelity models for the main classes of model truncations. Specifically, our neural closure models efficiently account for truncated modes in ROMs, capture the effects of subgrid-scale processes in coarse models, and augment the simplification of complex mathematical models.
We also provide adjoint equation derivations and network architectures needed to efficiently implement nDDEs, for both discrete and distributed delays. 
In the case of distributed delays, we propose a novel architecture consisting of two coupled neural networks, which eliminates the need for using recurrent architectures for incorporating memory. 
We find that our nDDE closures substantially improve nODE closures and outperform classic dynamic closures such as the Smagorinsky subgrid-scale model. We explain the better performance of nDDE closures based on information theory and the amount of past-information being included. 
Our first two classes of simulation experiments utilize the advecting shock problem governed by the Burger's partial differential equation (PDE), with low-fidelity models derived either by proper-orthogonal-decomposition Galerkin projection \cite{behzad2015sensitivity} or by reducing the spatial grid resolution.
Our third class of experiments considers 
marine biological models of varying complexities \cite{newberger2003analysis, fennel2014introduction,borja2014tales} and then their physical-biogeochemical extensions, 
with low-fidelity models obtained by aggregation of components and other simplifications of processes and parameterizations. Finally, we analyze computational complexity and explain the limited additional computational cost due to the presence of neural closure models.


\section{Closure Problems}
\label{sec: closure modeling}
The need for closure modeling in dynamical systems arises for a variety of reasons. They often involve computational costs considerations, but also include the lack of data to resolve complex real processes, the limited understanding of fundamental dynamics, and the inherent nonlinear growth of uncertainties due to model errors and predictability limits \cite[e.g.][]{robinson_et_al_OCEANS2002,lermusiaux_et_al_O2006a,lermusiaux_et_al_O2006b,robinson2017systematic}. In this section, we examine three main classes of low-fidelity models that can require closure modeling.

\subsection{Reduced Order Modeling}
\label{sec: ROM}
Let us consider a nonlinear dynamical system with state variable $u \in \mathbb{R}^N$ and the full-order-model (FOM) dynamics governed by,
\begin{equation}
    \frac{du(t)}{dt} = Lu(t) + h(u(t)), \quad \text{with} \quad u(0) = u_0 \, ,
\label{eq: generic dynamical system}
\end{equation}
where $L \in \mathbb{R}^{N \times N}$ is the linear, and $h(\cdot): \mathbb{R}^N \rightarrow \mathbb{R}^N$ the nonlinear, part of the system. We are mainly interested in dynamical systems whose solution could be effectively approximated on a manifold of lower dimension, $\mathcal{V} \in \mathbb{R}^m \subset \mathbb{R}^N$ \cite[e.g.][]{feppon_lermusiaux_SIMAX2018a}. Ideally, the dimension of this manifold is much smaller than that of the system, i.e.\;$m << N$. For the classic Galerkin-based reduced-order modeling, a linear decomposition of the form,
\begin{equation}
    u(t) \approx \bar{u} + V a
    \label{eq: decomposition}
\end{equation}
is used, where $\bar{u} \in \mathbb{R}^N$ is a reference value, the columns of $V = [v_1, ..., v_m] \in \mathbb{R}^{N \times m}$ a basis of the $m$-dimensional subspace $\mathcal{V}$, and $a\in \mathbb{R}^m$ the vector of coefficients corresponding to the reduced basis. A popular choice for this basis is the proper-orthogonal-decomposition (POD) that defines the subspace such that the manifold $\mathcal{V}$ preserves the variance of the system as much as possible when projected on $V$ for a given $m$. The reference value ($\bar{u}$) is then commonly chosen as the mean of the system state, in order to prevent the first reduced coefficient from containing the majority of the energy of the system and to help stabilize the reduced system \cite{holmes2012turbulence}. 

Now, substituting Eq.\;\ref{eq: decomposition} into Eq.\;\ref{eq: generic dynamical system}, and projecting the result on the orthonormal modes $V$, we obtain the following set of ordinary differential equations for the coefficients $a$,
\begin{equation}
    \frac{d a}{dt} = V^T L V a+ V^T h(\bar{u} + V a) + V^T L \bar{u} \, , \quad \text{with} \quad
    a (0) = V^T(u_0 - \bar{u}) \,.
    \label{eq: generic coefficient eqn}
\end{equation}
This $m$ dimensional system, with $m <<N$, is computationally much cheaper than the original FOM  Eq.\;\ref{eq: decomposition}. This method of dimensionality reduction is commonly referred to as the POD Galerkin Projection (POD-GP) method. It can suffer from a number of issues. First, the truncated modes can play an important role in the dynamical behaviour of the system, and neglecting them can thus lead to a very different forecast \cite{kutz_etal_DMD2016}. Second, the error in the reduced state may be simply too large for truncation, i.e.\;the POD reduction is not efficient. Third, if steady POD are employed, they may quickly become irrelevant for the evolving system state \cite{sapsis_lermusiaux_PHYSD2012,feppon_lermusiaux_SIMAX2018a,feppon_lermusiaux_SIREV2018}.
To address these issues, several methods try to represent the effect of the truncated modes. 
%
The most common approaches introduce a nonlinear parameterization of the coefficients \cite[e.g.][]{matthies2003nonlinear}
in Eq.\;\ref{eq: generic coefficient eqn}, however, they are not yet generally applicable to all classes of closures.
%

\begin{figure}[h!]
  \centering
  \includegraphics[width=.6\textwidth]{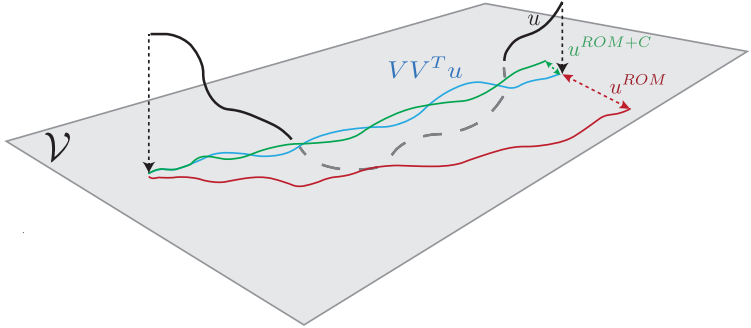}
  \caption{Geometric interpretation of the closure for reduced-order-models (ROMs). 
  $u$\;({\color{black}\rule[0.8mm]{0.5cm}{1pt}}): Solution to the full-order-model (FOM);~ $VV^Tu$\;({\color{blue}\rule[0.8mm]{0.5cm}{0.75pt}}): Projection of $u$ on the subspace $V$;~ 
  $u^{ROM}$\;({\color{red}\rule[0.8mm]{0.5cm}{0.75pt}}): Solution to the  proper-orthogonal-decomposition Galerkin-projection (POD-GP) ROM;~ and, 
  $u^{ROM+C}$\;({\color{green}\rule[0.8mm]{0.5cm}{0.75pt}}): Solution to POD-GP ROM with closure. Adapted from \cite{wang2020recurrent}.}
  \label{fig:ROM Closure}
\end{figure}

The geometric interpretation of the goal of closure modeling for ROMs is sketched in Fig.\;\ref{fig:ROM Closure}. The FOM solution of our dynamical system lies outside the lower dimension manifold, $\mathcal{V}$. A ROM approximate solution, denoted by $u^{ROM}$, starts with the projection of the full-order initial condition onto the manifold, $VV^Tu(0)$, but quickly diverges from the actual projection of the full-order solution onto the manifold ($VV^Tu$), often leading to a significant source of error. A closure model in this case basically attempts to keep the updated ``closed" solution, $u^{ROM+C}$, as close as possible to the actual projection of the FOM solution ($VV^Tu$) which can be seen as the truth.
\vskip -0.3truecm

\vskip -0.2truecm

\vskip -0.2truecm
\subsection{Subgrid-Scale Processes}
\label{sec: Subgrid-Scale Processes}
A key decision while setting up any numerical simulation is the selection of spatio-temporal resolution, which is in general limited by the computing power available. Using a coarse resolution (low-fidelity) model may however lead to a number of undesired artifacts, such as missing critical scales and processes for longer-term predictions or numerical diffusion that causes unintended or unacceptable results  \cite{yeung2018effects, laizet2015influence}. These artifacts become especially important in the case of ocean models. For example, present-day global observing systems and global model solutions only resolve open-ocean mesoscale processes ($\mathcal{O}(10-100km)$), but the submesocale (subgrid-scale) processes do have global consequences, in relation to the mechanisms of energy dissipation in the general circulation, vertical flux of material concentrations, and intermediate-scale horizontal dispersal of materials \cite{mcwilliams2016submesoscale,mcwilliams2017submesoscale,dauhajre2019nearshore}.
The neglected and unresolved scales along with their interactions with the resolved ones are then at the core of closure parameterizations. Most present oceanic models consist of a nonlinear system of PDEs, each of a nonlinear advection type, supplemented by other possible diagnostic nonlinear equations, and boundary conditions. There is however no unique way of defining such parameterizations and multiple approaches such as non-dimensional analyses, physical balance hypotheses, statistical correlation constraints, and other empirical methods are commonly employed to develop closure models. Similar statements can be made for atmospheric, Earth system, and climate models \cite{schneider2017earth}. For all these applications, a general approach for subgrid-scale closures would thus be most useful.

\subsection{Simplification of Complex Dynamical Systems}
\label{sec: Complexity of Models}
Due to incomplete understanding and limited measurements, it is common when modeling real dynamical systems in nature and engineering that the dynamics cannot be accurately explained just by using conservation laws and fundamental process equations. We refer to 
such systems as complex dynamical systems. The number of candidate models and equations can then be almost as large as the number of modelers.
The resulting models also  vary greatly in terms of their complexity. 
More complex models can capture key processes and feedbacks. Complexity is increased by adding more parameters and parameterizations to the existing components (state variables) of the dynamical model, but at some point, it quickly becomes inevitable to add and model new components to capture the underlying real processes accurately, hence further increasing model complexity \cite{nowak2006evolutionary,fennel2014introduction,may2019stability}.
This is common, for example, in marine ecosystem models, where simpler models only resolve the broad biogeochemical classes, while more complex models capture detailed sub-classes \cite{ward2013biogeochemical,los2010complexity}. Increasing the number of components however can come at great computational cost, can increase the overall uncertainty, and can lead to loss of accuracy or stability due to the nonlinearities.
Also, the unknown parameters for models with more components are calibrated from available data and the optimization process and parameter estimation quickly become challenging with the increase in complexity, due to the simultaneous explosion in the number of unknown parameters \cite{robinson_lermusiaux_Sea2002}. Thus, instead of adding more unknown parameterizations or increasing the number of components, one might use adaptive models \cite{lermusiaux_et_al_CS2004,lermusiaux_PhysD2007} and in general the present neural closure models with time delays to incorporate the effects of missing processes in low-complexity models, enabling them to adapt and emulate the response from high-complexity models.

\section{Theory and Methodology}
\label{sec: Methodology}
In this section, we develop the 
theory and methodology for learning data-assisted closure models for dynamical systems. We first review the Mori-Zwanzig (MZ) formulation \cite{stinis2015renormalized, chorin2000optimal, gouasmi2017priori} which derives the exact functional form of the effects of truncated dynamics for common reduced models. Unfortunately, apart for very simple linear dynamical systems, the use of this formulation is challenging without making unjustified approximations and simplifications. We then discuss the presence of delays in complex dynamical systems, and their impact on modeling \cite{glass2020nonlinear}. Motivated by both the MZ formulation and the presence of delays, we finally derive the new nDDEs and neural closure models, including adjoint equations and network architectures, for both discrete and distributed delays.

\subsection{Mori-Zwanzig Formulation and Delays in Complex Dynamical Systems}
\label{sec: MZ formulation}
Without loss of generality, the full nonlinear dynamical system model is written as,
\begin{equation}
        \frac{d u_k(t)}{dt} = R_k(u(t), t) \, , \quad \text{with}\quad u_k(0) = u_{0k}\,, \quad k \in \mathfrak{F} \,.
    \label{eq: very general dynamical system}
\end{equation}
The full state vector is  $u = (\{ u_k \}), \; k \in \mathfrak{F} = \mathfrak{R} \cup \mathfrak{U}$, where $\mathfrak{R}$ is the set corresponding to the resolved variables (e.g.\;coarse field or reduced variables), and $\mathfrak{U}$ the set corresponding to the unresolved variables  (e.g.\;subgrid field or complement variables), which as a union, $\mathfrak{F}$, form the set for full space of variables. We also denote $u = \{\hat{u}, \tilde{u}\}$ where $\hat{u} = (\{ u_k \}), \; k \in \mathfrak{R}$ and $\tilde{u} = (\{ u_k \}),\; k \in \mathfrak{U}$. Similarly, $u_0 = \{\hat{u}_0, \tilde{u}_0\} $, with $\hat{u}_0 = (\{ u_{0k} \}), \; \in \mathfrak{R}$ and $\tilde{u}_0 = (\{ u_{0k} \}), \;  k \in \mathfrak{U}$.

The Mori-Zwanzig (MZ) formulation allows rewriting the above nonlinear system of ODEs as,
\begin{equation}
    \begin{split}
        \frac{\partial }{\partial t}u_k(u_0, t) = \underbrace{R_k(\hat{u}(u_0, t))}_{\text{Markovian}} + \underbrace{F_k(u_0, t)}_{\text{Noise}} + \underbrace{\int_0^t K_k(\hat{u}(u_0, t - s), s) ds}_{\text{Memory}} , \quad k \in \mathfrak{R}\,,
    \end{split}
    \label{eq: rewrite linear PDE system in MZF concise}
\end{equation}
where $R_k$ is as in the full model dynamics  (Eq.\;\ref{eq: very general dynamical system}).
Importantly, the above equation is an exact representation of Eq.\;\ref{eq: very general dynamical system} for the resolved components. A derivation is provided in the \textit{Supplementary Information} (Sec.\;\ref{sec: MZ formulation in supp info}).
%
Eq.\;\ref{eq: rewrite linear PDE system in MZF concise} provides useful guidance for closure modeling. The first term in Eq.\;\ref{eq: rewrite linear PDE system in MZF concise} is the Markovian term dependent only on the values of the variables at the present time, while the closure consists of two terms: the noise term and a memory term that is non-Markovian. 
We can further simplify Eq.\;\ref{eq: rewrite linear PDE system in MZF concise} by applying the $P$ projection and using the fact that the noise term lives in the null space of $P$ for all times, which could be easily proved. For ROMs with initial conditions devoid of any unresolved dynamics, i.e.\;$\tilde{u}_0 = 0$ and thus $u_0 = \hat{u}_0$, we then retain the exact dynamics after the projection step, noticing in this case that $Pu_k(u_0, t) = u_k(\hat{u}_0, t), \forall k \in \mathfrak{R}$,
\begin{equation}
    \begin{split}
        \frac{\partial }{\partial t}u_k(\hat{u}_0, t) = P R_k(\hat{u}(\hat{u}_0, t)) + P\int_0^t K_k(\hat{u}(\hat{u}_0, t - s), s) ds , \quad k \in \mathfrak{R}\,.
    \end{split}
    \label{eq: MZF with no unresolved ICs}
\end{equation}
Hence, for such systems, the closure model would only consider the non-Markovian memory term. The above derivation of the MZ formulation has been adapted from \cite{stinis2015renormalized,wang2020recurrent,gouasmi2017priori}.

The MZ formulation clearly shows that a non-Markovian closure term requiring time-lagged state information is theoretically needed to model the unresolved or missing dynamics. This theoretical basis directly applies to the first two classes of low-fidelity dynamical systems (Sec.\;\ref{sec: closure modeling}), ROMs and coarse resolution models. 
For the third category, the simplification of complex dynamical systems, we emphasize biological and chemical systems. Many are modeled using ODEs, with one state variable per biological or chemical component. Such ODEs implicitly assume that information between state variables is exchanged instantaneously.
In reality, however, there are often time-delays for several reasons.
%
First, changes in populations or reactions have non-negligible time-scales \cite[e.g.][]{kuang1993delay,DellAnna2020, otto2019nonlinear}. Such time-scales are introduced in more complex models by modeling intermediate state variables. Hence, the time response of lower-complexity models can be comparable to that of high-complexity models only by explicitly introducing delays \cite{kuang1993delay,bocharov2000numerical,glass2020nonlinear,tokuda2019reducing}.
Second, many reactive systems are modeled assuming smooth concentration fields of state variables governed by PDEs with fluid flow advection and/or mixing, leading to advection-diffusion-reaction PDEs
\cite{hundsdorfer2013numerical,faugeras2007modeling}. In that case, simplified models still require time-delays due to the neglected reactive or biogeochemical dynamics but now also due to truncated modes and/or subgrid-scale processes of numerical models.
For all of these reasons, the need for memory based closure terms is clearly justified to represent complex dynamical systems.

There are some results for data-assisted\;/\;data-driven closure modeling based on the MZ formulation. Some schemes create a coupled system of stochastic differential equation using appropriate hidden-variables for approximate Markovization of the non-Markovian term \cite{kondrashov2015data, boers2017inverse}. Others use a variational approach to derive nonlinear parameterizations approximating the Markovian term \cite{chekroun2019variational}.
Schemes using machine learning to learn non-Markovian residual of the high- and low-fidelity dynamics limit themselves to specific functional forms for the residual term, simple Euler time-stepping scheme, and very frequent
and uniformly-spaced
training data \cite{pan2018data, wang2020recurrent,wan2018data}. They also lack the rigorous use of the theory for time-delay systems \cite{richard2003time}.




\subsection{Neural Delay Differential Equations}
\label{sec: nDDE}

The non-Markovian closure terms with time-lagged state information lead us to delay differential equations (DDEs) \cite{diekmann2012delay}. DDEs have been widely used in many fields such as biology \cite{smith2011introduction,macdonald2008biological}, pharmacokinetic-pharmacodynamics \cite{koch2014modeling}, chemistry \cite{roussel1996use}, economics \cite{keller2010generalized}, transportation \cite{matsuya2015exact}, control theory \cite{kunisch1982approximation}, climate dynamics \cite{ghil2015collection, bhattacharya1982internal}, etc. Next, we summarize the state of the art for learning and solving differential equations using neural-networks (NNs) and develop theory and schemes for neural DDEs including adjoint equations for backpropagation.


The interpretation of residual networks as time integration schemes and flow maps for dynamical systems has 
led to pioneering development of neural ordinary differential equations (nODEs) \cite{chen2018neural}. A nODE parameterizes an ODE using a neural-network and solves the initial value problem (IVP) given by,
\begin{equation}
    \frac{du(t)}{dt} = f_{NN}(u(t), t; \theta) \, , \quad t \in (0, T]\,, \quad \text{with}\quad
    u(0) = u_o \, ,
    \label{eqn: nODE}
\end{equation}
where $f_{NN}$ is the prescribed neural-network and $\theta$ are the weights. Starting from the initial conditions, the nODE (Eq.\;\ref{eqn: nODE}) is integrated forward in time using any time-integration scheme, and then gradients are computed based on a loss function using the adjoint sensitivity method. The gradient computation boils down to solving a second ODE backwards in time. Using standard backpropagation for Eq.\;\ref{eqn: nODE} has however several issues: it would be very memory expensive as one needs to store the state at every time step; its computational cost would increase when using higher-order time-integration or \textit{implicit} schemes; and, it might become infeasible if the forward time-integration code does not support automatic differentiation. The adjoint method, however, provides a backpropagation for nODEs \cite{chen2018neural} that is memory efficient and flexible as it treats the time-integration scheme as a "black-box". In our case, we need to incorporate state-delays. Though  extending the nODE framework to incorporate DDEs comes under the ambit of universal differential equations (UDEs) \cite{rackauckas2020universal}, deeper investigations are warranted. First, the UDEs are presently implemented using the Julia library DiffEqFlux.jl \cite{rackauckas2019diffeqflux} which can perform automatic adjoint equation solves, but other popular open-source languages such as Python and R, and ML-Frameworks such as TensorFlow \cite{tensorflow2015-whitepaper}, PyTorch \cite{paszke2019pytorch}, etc., would require explicit derivation and coding of the corresponding adjoint equations. 
Second, we need to study two different types of DDEs, the discrete and distributed delays, which it turns out require different architectures. Next, we thus develop the theory and schemes for efficient implementation of neural delay differential equations (nDDEs) in any programming language.

\subsubsection{Discrete Delays}
\label{sec: nDDE-RNN}
The most popular form of delay differential equations (DDEs) is,
\begin{equation}
\begin{split}
\frac{du(t)}{dt} = f(u(t), u(t - \tau_1), ..., u(t - \tau_K), t), \quad t \in (0, T]\,, \quad\text{with} \quad
u(t) = h(t), \quad t \leq 0\,,
\end{split}
\end{equation}
where $\tau_1, ..., \tau_K$ are $K$ number of discrete-delays (discrete DDEs). Instead of a single initial value as in the case of ODEs, DDEs require specification of a history function, $h(t)$. 
Due to the presence of a given fixed number delays, we can parameterize the above system by replacing the time-derivative function with potentially any type of NNs. For example, to use fully-connected NNs we would concatenate all the delayed states vertically to form the input vector, or concatenate them horizontally to form an input matrix for a convolutional NN. However, recurrent NN (RNN) architectures, such as simple-RNNs, LSTMs, GRUs, etc., are ideal and most efficient for our need due to the time-series nature of the delayed states. We can assume that the discrete delays are evenly spaced (this is not a hard requirement as we can easily extend schemes to irregularly spaced discrete-delays using ODE-RNNs \cite{rubanova2019latent}, but for brevity we make this assumption) and use a RNN with weights $\theta$. 
Hence, our new discrete-DDE system can be written as,
\begin{equation}
\begin{split}
\frac{du(t)}{dt} = f_{RNN}(u(t), u(t - \tau_1), ..., u(t - \tau_K), t; \theta), \quad t \in (0, T]\,, \;\; \text{with} \;\;
u(t) = h(t), \quad t \leq 0 \, ,
\end{split}
\label{eq: DDE-RNN main equation}
\end{equation}
where $f_{RNN}(\bullet; \theta)$ is the recurrent architecture. We refer to this parameterization of discrete DDEs as \textit{discrete-nDDE}. 
%
%
The graphical representation of Eq.\;\ref{eq: DDE-RNN main equation} in time-discretized form is depicted in Figure \ref{fig:nDDDE}. 
Let data be available at $M$ times, $T_1 < ... <T_M \leq T$. 
%
We then optimize the total loss function given by, $\mathcal{L} = \int_0^T \sum_{i=1}^M l(u(t))\delta(t - T_i)dt$, where $l(\bullet)$ are scalar loss functions such as mean-squared-error (MSE), and $\delta(t)$ is the Kronecker delta function.
%
To perform this optimization with any gradient descent algorithm, we need the gradient of the loss function w.r.t. the weights of the RNN, $\theta$.
Using the adjoint sensitivity method \cite{calver2017numerical} to compute the required gradients, we start by writing the Lagrangian for the above system, 
\begin{equation}
\begin{split}
   L =& \mathcal{L}(u(t)) + \int_0^T \lambda^T(t) \left( d_t{u(t)} - f_{RNN}(u(t), u(t-\tau_1), ..., u(t-\tau_K), t; \theta)\right)dt \\
   & + \int_{-\tau_K}^0 \mu^T(t)(u(t) - h(t))dt \, ,
   \end{split}
\end{equation}
where $\lambda(t)$ and $\mu(t)$ are the Lagrangian variables. 
%
In order to find the gradients of $L$ w.r.t.~$\theta$, we first solve the following adjoint equation (for brevity we denote, $\frac{\partial}{\partial (\bullet)} \equiv \partial_{(\bullet)}$ and $\frac{d}{d (\bullet)} \equiv d_{(\bullet)}$),
\begin{eqnarray}
   & d_t\lm^T(t) = \sum_{i = 1}^M\partial_{u(t)} l(u(t))\delta(t - T_i) - \lm^T(t)\partial_{u(t)}f_{RNN}(u(t), u(t - \tau_1), ..., u(t - \tau_K), t; \theta) \nonumber \\
   & - \sum_{i = 1}^K \lm^T(t+\tau_i)\partial_{u(t)}f_{ RNN}\left(u(t+\tau_i), u(t-\tau_1+\tau_i), ..., u(t-\tau_K+\tau_i), t + \tau_i; \theta \right), \; t \in [0, T) \nonumber \\
   & \lm(t) = 0, \qquad t\geq T \;. \qquad\qquad\qquad\qquad\qquad\qquad\qquad\qquad\qquad\qquad\qquad\qquad\qquad\qquad
   \label{eq: adjoint nDDE-RNN}
\end{eqnarray}
Details of the derivation of the above adjoint Eq.\;\ref{eq: adjoint nDDE-RNN} are in the accompanying \textit{Supplementary Information}. Note that Eq.\;\ref{eq: adjoint nDDE-RNN} needs to be solved backward in time, and one would require access to $u(t), ~0\leq t \leq T$. In the original nODE work \cite{chen2018neural},  Eq.\;\ref{eq: DDE-RNN main equation} is solved backward in time and augmented with the adjoint Eq.\;\ref{eq: adjoint nDDE-RNN}, so as to shrink the memory footprint by avoiding the need to save $u$ at every time-step. Solving Eq.\;\ref{eq: DDE-RNN main equation} backward can however lead to catastrophic numerical instabilities as is well known in data assimilation \cite{wunsch1996ocean,robinson_et_al_Sea1998}. 
Improvements have been proposed, such as the ANODE method \cite{gholami2019anode}, but they are not applicable in case of DDEs.
In our present implementation, in order to access $u(t), ~0\leq t \leq T$, while solving the adjoint equation, we create and continuously update an interpolation function using the $u$ obtained at every 
time-step as we solve Eq.\;\ref{eq: DDE-RNN main equation} forward in time.
%
To be more memory efficient, we can, for example, use the method of \textit{checkpointing} \cite{griewank1992achieving}, or the interpolated reverse dynamic method (IRDM) \cite{daulbaev2020interpolation}. 
After solving for $\lambda$, we can compute the required gradients as,
\begin{equation}
   \dt L = - \int_0^T \lambda^T(t)\partial_{\theta}f_{RNN}(u(t), u(t - \tau_1), ..., u(t - \tau_K), t; \theta)dt \;.
\label{eqn: dL_dtheta discrete DDE}
\end{equation} 
Finally, using any gradient descent algorithm, we can find the optimal values of the weights $\theta$.

\subsubsection{Distributed Delays}
\begin{figure}
  \centering
  \subfloat[][Discrete-nDDE]{\includegraphics[width=.45\textwidth]{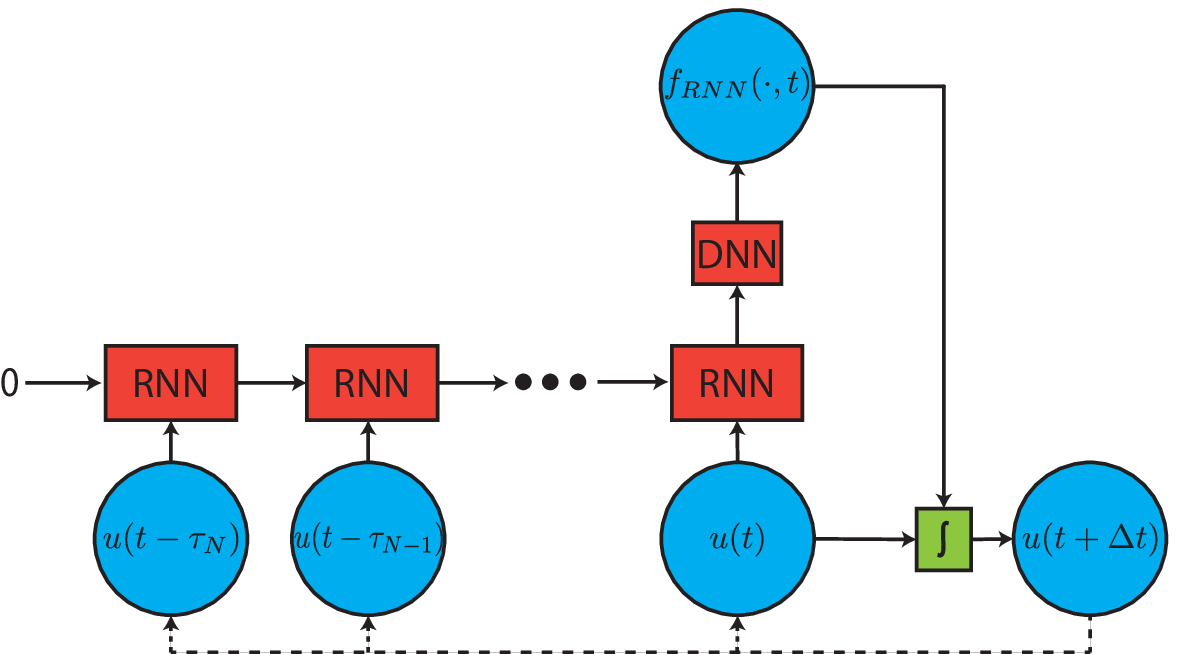}\label{fig:nDDDE}} \qquad
\subfloat[][Distributed-nDDE]{\includegraphics[width=.45\textwidth]{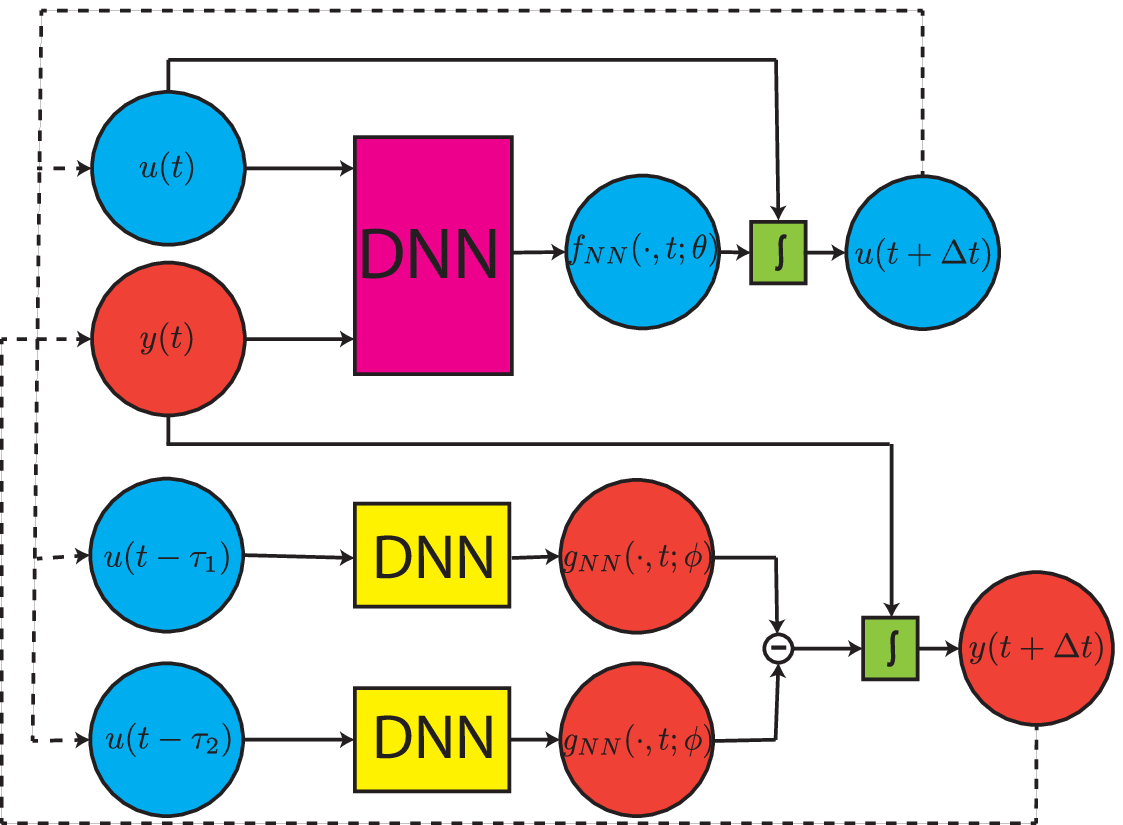}\label{fig:nDistDDE}}
  \caption{Graphical representation of the time discretized neural delay differential equations (nDDEs). The blocks labeled \textit{RNN} and \textit{DNN} represent any recurrent or deep neural-network architectures respectively. The block labeled $\int$ symbolizes any time-integration scheme.
  }
  \label{fig:DDE architecture}
\end{figure}

In some applications, the delay is distributed over some past time-period \cite{rasmussen2003analysis},
\begin{equation}
\begin{split}
\frac{du(t)}{dt} = f\left(u(t), \int_{t-\tau_2}^{t-\tau_1} g(u(\tau), \tau)d\tau, t \right), \quad t \in (0, T]\,, \quad \text{with} \quad u(t) = h(t), \quad t \leq 0 \, .
\end{split}
\label{eqn: distributed-DDE}
\end{equation}
It should be noted that the discrete DDEs can be written as a special case of distributed DDEs using dirac-delta functions. We can approximate the two functions $f$ and $g$ using two different neural-networks, and re-write the above Eqs.\;\ref{eqn: distributed-DDE} as our new coupled discrete DDEs,
\begin{equation}
\begin{split}
\frac{du(t)}{dt} &= f_{NN}\left(u(t), y(t), t; \theta \right), \quad t \in (0, T] \\
\frac{dy(t)}{dt} &= g_{NN}(u(t-\tau_1), t-\tau_1; \phi) - g_{NN}(u(t-\tau_2), t-\tau_2; \phi), \quad t \in (0, T] \\
\text{with}\quad u(t) &= h(t), \quad \tau_2 \leq t \leq 0 \, , \quad\text{and}\quad y(0)  = \int_{-\tau_2}^{-\tau_1} g_{NN}(h(t), t; \phi)dt \, ,
\end{split}
\label{eqn: distributed-nDDE}
\end{equation}
where $f_{NN}(\bullet;\theta)$ and $g_{NN}(\bullet; \phi)$ are the two NNs parameterized by $\theta$ and $\phi$ respectively. We refer to this parameterization of distributed DDEs as \textit{distributed-nDDE}. The graphical representation of the above system (Eqs.\;\ref{eqn: distributed-nDDE}) in time-discretized form is depicted in Figure \ref{fig:nDistDDE}. 
%
%
Interestingly in the case of distributed-delays, we obtain a novel architecture consisting of two coupled NNs, which enables us to incorporate memory without the use of any recurrent networks such as RNN, LSTMs, GRUs, etc.
We can consider $f_{NN}$ as the main network, and $g_{NN}$ as the auxiliary network. Again, we define a scalar loss function given by $\mathcal{L} = \int_0^T \sum_{i=1}^M l(u(t))\delta(t - T_i)dt$ for the available data at $M$ times, $T_1 < ... <T_M \leq T$. The Lagrangian for the above system is,
\begin{equation}
\begin{split}
   L =& \mathcal{L}(u(t)) + \int_0^T \lm^T(t) (d_t u(t) - f_{NN}(u(t), y(t), t; \theta))\, dt \\
   & + \int_0^T \mu^T(t) \left({ d_t y(t) - g_{NN}(u(t - \tau_1), t-\tau_1; \phi)  + g_{NN}(u(t - \tau_2), t-\tau_2; \phi) }\right)dt  \\
   & + \int_{-\tau_2}^0 \gamma^T(t)(u(t) - h(t) ) dt  + \alpha^T \left (y(0) - \int_{-\tau_2}^{-\tau_1} g_{NN}(h(t), t; \phi)dt \right) \,, 
\end{split}
\end{equation}
where $\lambda(t), \mu(t), \gamma(t)$, and $\alpha$ are the Lagrangian variables. In order to find the gradients of $L$ w.r.t.~the parameters of the two NNs, we first solve the following coupled adjoint equations backward in time,
\begin{equation}
\begin{split}
   d_t\lm^T(t) =& \sum_{i = 1}^M\partial_{u(t)} l(u(t))\delta(t - T_i) - \lm^T(t)\partial_{u(t)}f_{NN}(u(t), y(t), t; \theta) \\
   & - \mu^T(t+\tau_1) \partial_{u(t)}g_{NN}(u(t), t; \phi)  +  \mu^T(t+\tau_2) \partial_{u(t)}g_{NN}(u(t), t; \phi)\,, \; t \in [0, T) \\
   d_t\mu^T(t) =& - \lm^T(t)\partial_{y(t)}f_{NN}(u(t), y(t), t; \theta)\,, \qquad t \in [0, T) \\
   \lm^T(t)  =& ~ 0 \quad \text{and} \quad \mu^T(t) = 0, \qquad t \geq T \, .
\end{split}
\label{eq: adjoint nDDE-Dist} 
\end{equation}
Details of the derivation of the above adjoint Eq.\;\ref{eq: adjoint nDDE-Dist} are in the \textit{Supplementary Information}. For accessing $u$ values while solving the adjoint equations, we use the same approach as for our discrete-nDDE (Sec.\;\ref{sec: nDDE-RNN}). After solving for $\lambda$ and $\mu$, we can compute the required gradients as,
\begin{equation}
\begin{split}
\dt L =& - \int_0^T \lambda^T(t)\partial_{\theta}f_{NN}(u(t), y(t), t; \theta)dt \, , \\ 
\di L =& - \int_0^T \mu^T(t) \left( \partial_{\phi}g_{NN}(u(t - \tau_1), t-\tau_1; \phi) - \partial_{\phi}g_{NN}(u(t-\tau_2), t-\tau_2; \phi)\right) dt \\
   & - \mu^T(0)\int_{-\tau_2}^{-\tau_1} \partial_{\phi} g_{NN}(h(t), t; \phi)dt \,.
\end{split}
\label{eqn: dL_dtheta dL_dphi distributed DDE}
\end{equation}
Finally, using any gradient descent algorithm, we can optimize the neural-networks $f_{NN}$ and $g_{NN}$, and find the optimal values of the weights $\theta$ and $\phi$.

\subsection{Neural Closure Models}
Now that we have the framework for representing delay differential equations using neural-networks, we can replace the non-Markovian memory term in Eq.\;\ref{eq: MZF with no unresolved ICs} using nDDEs to obtain a hybrid closure model which could be trained using data from high-fidelity simulations or real observations. The modified low-fidelity dynamical system with the nDDE closures, which approximates the high-fidelity model would be given by,
\begin{equation}
    \begin{split}
        \frac{\partial \hat{u}(t)}{\partial t} &= \underbrace{PR(\hat{u}(t))}_{\text{Low-Fidelity}} + \underbrace{f_{RNN}(\hat{u}(t), \hat{u}(t - \tau_1), ..., \hat{u}(t - \tau_K), t; \theta)}_{\text{Neural Closure}} \\
\hat{u}(0) &= \hat{u}_0, \qquad t\leq 0
    \end{split}
    \label{eq: low-fid with discrete DDE}
\end{equation}
using discrete-delays, or by,
\begin{equation}
\begin{split}
        \frac{\partial \hat{u}(t)}{\partial t} &= \underbrace{PR(\hat{u}(t))}_{\text{Low-Fidelity}} + \underbrace{f_{NN}\left(\hat{u}(t), \int_{t-\tau_2}^{t-\tau_1} g_{NN}(\hat{u}(\tau), \tau; \phi)d\tau, t; \theta \right)}_{\text{Neural Closure}} \\
\hat{u}(0) &= \hat{u}_0, \qquad t\leq 0
    \end{split}
    \label{eq: low-fid with distributed DDE}
\end{equation}
using distributed-delays. The initial conditions at $t=0$, $\hat{u}_0$, can be used for $t < 0$ as well, as an approximation. Apart from the neural-network architectures, the amount of delay to be used also becomes a hyperparameter to tune. These novel \emph{neural closure models} provides extreme flexibility in designing the non-Markovian memory term in order to incorporate subject matter expert insights. At the same time, we can also learn the unknown parts of the Markovian low-fidelity model using nODEs if the need arises. Next, we will compare the performance and advantages of using no-delays (nODEs), discrete-delays (discrete-nDDEs), and distributed-delays (distributed-nDDEs) in closure terms for various low-fidelity dynamical systems.



\section{Application Results and Discussion}
\label{sec: results and discussions}

After presenting the main classes of low-fidelity dynamical models that require closure (Sec.\;\ref{sec: closure modeling}),
we derived a novel, versatile, and rigorous methodology for learning and modeling non-Markovian closure terms using nDDEs
(Sec.\ \ref{sec: Methodology}). The resulting neural closure models have their underpinning in the Mori-Zwanzig formulation and the presence of inherent delays in models of complex dynamical systems such as biogeochemical systems. Now, we evaluate the performance and advantages of these new neural closure models over those of neural ODEs (Markovian).

We run experiments encompassing each of the classes of low-fidelity models (Sec.\;\ref{sec: closure modeling}).
For each experiment,
we follow the same training protocol for nODEs (no-delays) and the two nDDEs, discrete-nDDE (discrete-delays) and distributed-nDDE (distributed-delays), closure models.
The training data are regularly sampled from high-fidelity simulations in all experiments, but this is not a requirement. 
%
We use performance over the validation period (past the period for which high-fidelity data snapshots are used for training) to fine tune various training related hyperparameters. 
The final evaluation is based on much longer-term future prediction performance, well past these periods.
%
As the field of scientific machine learning (SciML; \cite{SciMLwebsite}) is relatively new, the metrics for performance evaluation vary greatly. On the one hand, many learning studies randomly sample small time-sequences 
from a given period for which high-fidelity data are available, and then split them into training, validation, and test sets \cite{maulik2020time}. 
As the training, validation, and test sets belong to the same time-domain, hence, the learned networks are only evaluated for their interpolation performance and predicting the unseen data becomes easy for them.
On the other hand, for the few studies where the training and test (prediction) periods do not overlap, the prediction period is often much shorter than the training period \cite{yuval2020use}. 
In the present work, we consider a more stringent evaluation. 
First, our validation period does not overlap the training period. Second,
our future prediction period is equal to or much longer than the training and validation periods, and has no overlap with either.  Hence, we strictly measure the out-of-sample/generalization performance of the learned network for its extrapolation capabilities into the future.
Of course, other evaluation metrics are possible and there is indeed a need for standardization of evaluation procedures in the SciML community. 
In the rest of the paper, for all the figure, table, and section references prefixed with ``SI-", we direct the reader to the \textit{Supplementary Information}.

\subsection{Experiments 1: Advecting Shock - Reduced Order Model}
\label{sec: exp1}
\begin{figure}
  \centering
  \includegraphics[width=1\textwidth]{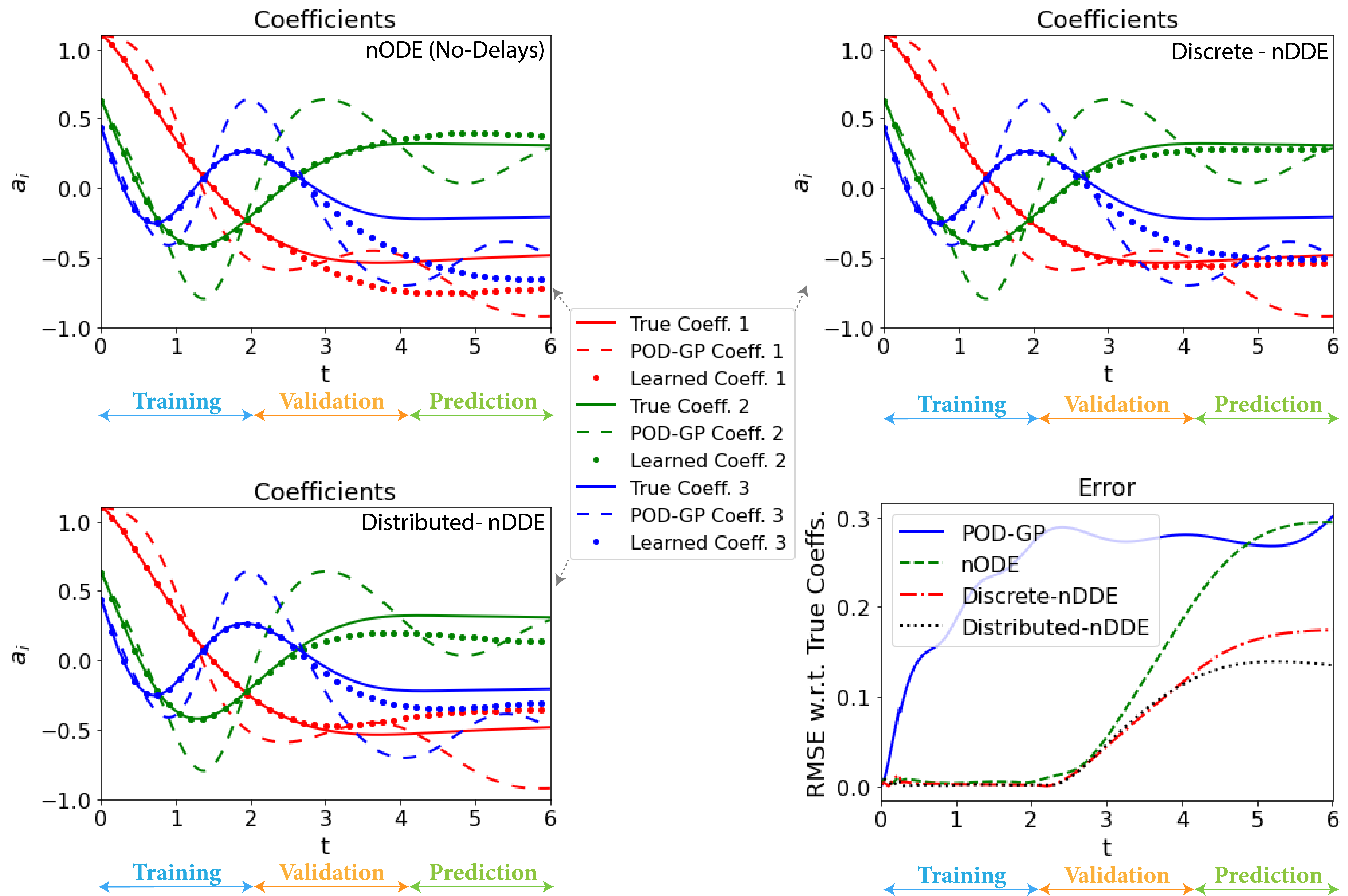}
  \caption{
  Comparison of the true coefficients (\textit{solid}) with the coefficients from the POD-GP ROM (\textit{dashed-dot}) and from
  the POD-GP ROMs augmented with the three different learned neural closure models at the end of training (\textit{dashed}). 
  For each neural closure, the training period is from $t=0~\text{to}~2.0$, the validation period from $t=2.0~\text{to}~4.0$, and the future prediction period from $t=4.0~\text{to}~6.0$.
  \textit{Top-left:} neural ODEs with no-delays (nODE); \textit{Top-right:} neural DDEs with discrete-delays (Discrete-nDDE); \textit{Bottom-left:} neural DDEs with distributed-delays (Distributed-nDDE). \textit{Bottom-right:} Evolution of root-mean-squared-error $(\text{RMSE}(t) = \sqrt{\frac{1}{3}\sum_{k=1}^3 |a_k^{pred}(t) - a_k^{true}(t)|^2})$ of coefficients from the four different ROMs. These results correspond to the architectures detailed in Table \ref{table: Exp1 and 2 architecture}.
  }
  \label{fig: Exp1_Comparison_Coeff_Plots}
\end{figure}

For the first experiments, neural closure models learn the closure of proper-orthogonal-decomposition Galerkin projection (POD-GP) based reduced order model of the advecting shock problem. The full-order-model (FOM) for this problem is given by the Burger's equation,
\begin{equation}
\label{eq: Advecting shock}
\frac{\partial u}{\partial t} + u\frac{\partial u}{\partial x} = \nu \frac{\partial^2 u}{\partial^2 x} \,,
\end{equation}
where $\nu$ is the non-dimensional diffusion coefficient. The initial and boundary conditions are
\begin{equation}
\label{eq: Advecting shock ICs and BCs}
\begin{split}
   u(x, 0) &= \frac{x}{1 + \sqrt{\frac{1}{t_0}} \exp\left(Re\frac{x^2}{4} \right)}\,, \quad  u(0, t) = 0, \quad \text{and} \quad u(L, t) = 0 \,,
   \end{split}
\end{equation}
where $Re = 1/\nu$ and $t_0 = \exp(Re/8)$. Let the POD of the state variable $u(x, t)$ be given by, $u(x, t) = \bar{u}(x) + \sum_{i=1}^m u_i(x) a_i(t)$, we obtain the reduced-order equations as outlined in Section \ref{sec: ROM},
\begin{equation}
\label{eq: advecting shock ROM}
\begin{split}
  \frac{d a_k}{dt} = & - \left\langle \bar{u} \frac{\partial \bar{u}}{\partial x}, u_k\right\rangle -
  a_i \left\langle u_i \frac{\partial \bar{u}}{\partial x}, u_k\right\rangle - 
  a_j \left\langle \bar{u} \frac{\partial u_j}{\partial x}, u_k\right\rangle  - 
  a_i a_j\left\langle u_i \frac{\partial u_j}{\partial x}, u_k\right\rangle \\ &
  + \left\langle \nu \frac{\partial^2 \bar{u}}{\partial x^2}, u_k\right\rangle +
  a_i \left\langle \nu \frac{\partial^2 u_i}{\partial x^2}, u_k\right\rangle  \, ,\quad \text{with}\quad
  a_k(0) =  \langle (u (x, 0) - \bar{u}(x)), u_k(x)  \rangle \, .
  \end{split}
\end{equation}
We solve the FOM (Eqs.\;\ref{eq: Advecting shock} and \ref{eq: Advecting shock ICs and BCs}) for $Re=1000$, $L=1$, and maximum time $T = 4.0$. The singular value decomposition (SVD) of this solution form the POD modes for the ROM.
We only keep the first three modes which capture $60.8\%$ of energy, and evolve the corresponding coefficients using Eq.\;\ref{eq: advecting shock ROM}, thus requiring a closure. 
The high-fidelity or true coefficients are obtained solving the FOM (Eq.\;\ref{eq: Advecting shock}) with initial conditions without the contribution from the unresolved modes, i.e. $u(x, 0) = \bar{u}(x) + \sum_{i=1}^3 u_i(x)a_i(0)$, and projecting the obtained solution onto the first three modes. For comparison, we also present the true coefficients in Fig. \ref{fig: Exp1_Comparison_Coeff_Plots}, which is what the ROMs with neural closure are trying to match.
%
For this true data generation, we solve the FOM using an explicit Runge-Kutta (RK) time-integration of order (4)5 
(\textit{dopri5}; \cite{hairer1993solving}) with adaptive time-stepping (storing data at time-steps of $\Delta t = 0.01$) and grid spacing of $\Delta x = 0.01$, using finite-difference schemes (upwind for advection and central difference for diffusion).

Our three test periods for the advecting shock ROM (Eq.\;\ref{eq: advecting shock ROM}) with three modes are as follows.
For training our neural closure models, we only use the true coefficient values up to time $t=2.0$.
For validation (used only to tune hyperparameters), we use true coefficient values from $t=2.0$ to $t=4.0$.
Finally for testing, we make a future prediction from $t=4.0$ to final time $T = 6.0$.
We  compare the three different closures: nODE (no-delays), discrete-nDDE, and distributed-nDDE with architecture details presented in Table \ref{table: Exp1 and 2 architecture}. 
The architectures are not exactly the same for the three cases, but they are set-up to be of comparable expressive power. Mostly, we employ a bigger architecture for the no-delays case in order to help it compensate the lack of past information. We also ensure that the networks are neither under-parameterized nor over-parameterized. 
Along with the classical hyperparameters such as batch size, number of iterations per epoch, number of epochs, learning rate schedule, etc., we also have the delay values ($\tau_1, ..., \tau_K$ for discrete-nDDE; and $\tau_1, ~\tau_2$ for distributed-nDDE) as additional hyperparameters to tune. 
We chose to 
use six discrete delays
for the discrete-nDDE in the present experiments. 
%
The values of other hyperparameters are given in Sec.\;\ref{SI: hyperparameters}.
%
For evaluation, at each epoch, we evolve the coefficient of the learned system $(\{a^{pred}(T_i) = \{a_k^{pred}(T_i)\}_{k=1}^3\}_{i=1}^M)$ using the RK time-integration scheme mentioned earlier, and compare them with the true coefficients $(\{a^{true}(T_i)\}_{i=1}^M)$ using the time-averaged $L_2$ error, $\mathcal{L} = \allowdisplaybreaks  \frac{1}{M} \sum_{i=1}^M\left(\sqrt{\sum_{k=1}^3 |a_k^{pred}(T_i) - a_k^{true}(T_i)|^2}\right)$, which is also our loss function for training.
The error for the time period $t=0~\text{to}~2.0$ forms the training loss, the error for $t=2.0~\text{to}~4.0$ the validation loss, and the error for $t=4.0~\text{to}~6.0$ the prediction loss.

The performance of the three neural closure models after 200 epochs (the stochastic gradient descent nearly converges, see Fig.\;\ref{fig: Exp_1_train_val_loss_all}) is evaluated by comparison with the true coefficients and with the POD-GP coefficients 
spanning training, validation, and future prediction periods.
Results are shown in Figure \ref{fig: Exp1_Comparison_Coeff_Plots}. 
The details of the architectures employed are in Table \ref{table: Exp1 and 2 architecture}. We find that using no-delays (nODE), discrete-delays (discrete-nDDE), and distributed-delays (distributed-nDDE) perform equally well for the training period, exactly matching the true coefficients. 
As soon as one enters the validation period, all the neural closure models starts to slightly diverge, with the nODE diverging the most by the end of prediction period. 
Importantly, both nDDE closures maintain a great improvement over just using the POD-GP model, and showcase a better performance than the nODE closure, even though the latter had a deeper architecture with significantly more trainable parameters.
We also find that the performance of the distributed-nDDE closure is a little better than that of the discrete-nDDE closure for the prediction period.
%
In a similar set of experiments, Maulik et. al., 2020 (section 3.1, ``Advecting Shock", \cite{maulik2020time}) used nODE and LSTM to learn the time evolution of the first three high-fidelity (true) coefficients without utilizing the known physics/low-fidelity model. As a result, they required bigger architectures and more time samples for training data than we do. This confirms our benefits of learning only the unknown closure model.
Due to the highly nonlinear nature of neural networks, analytical stability analyses are not direct. 
Nonetheless, we provide empirical stability results by reporting the evolution of the root-mean-square-error (RMSE)  (Fig.\;\ref{fig: Exp1_Comparison_Coeff_Plots}).
We find that both discrete-nDDE and distributed-nDDE closures, due to the existence of delays, may have a stronger dissipative character and thus show much better stability at later times than the POD-GP and the nODE closure.
%

One might expect the distributed-delay (distributed-nDDE) to always perform better than the discrete-nDDE closure because of the presence of the integral of the state variable over a delay period instead of the state variable at specific points in the past. The former thus seemingly contains more information, but there is in fact no guarantee for this being true in all cases. 
We can derive an intuition for this from information theory. According to the data processing inequality \cite{cover1999elements}, let $X$ and $Y$ be two random variables, then,
\begin{equation}
    I(g(X); Y) \leq I(X; Y) \, ,
\end{equation}
where $I$ is the mutual information and $g$ is any function which post-processes $X$. Now, if $X$ is composed of $K$ random variables, $X = \{X_1, ..., X_K\}$, and $g(X) = X_1 + ... + X_K$, then,
\begin{equation}
\label{eq: data-processing-ineq-sum}
    I(X_1 + ... + X_K; Y) \leq I(\{X_1, ..., X_K\}; Y) \, .
\end{equation}
If we consider the effect of the integral of the state variable over the delay period in the case of distributed-nDDE as a data processing step, this might actually be decreasing the information content as compared to the discrete-nDDE closure. We use ``might", even though Eq.\;\ref{eq: data-processing-ineq-sum} is a strong bound, because in the present experiments we only use 
six 
delay values for the discrete-nDDE, while the integral in the distributed-nDDE is computed using many past state values, and also the architectures are different. Hence, a direct comparison using the data processing inequality (Eq.\;\ref{eq: data-processing-ineq-sum}) is not possible, but it provides us with a plausible explanation.

In addition to the results just illustrated, we completed many other experiments-1 to assess the sensitivity of our framework to various hyperparameters. 
In all cases, the time-period corresponding to the training data should be at least equal to one characteristic time-scale of the dynamics, otherwise the prediction performance deteriorated, 
as shown in the Fig.\;\ref{fig: Exp1_RMSE_with_time_train_time} and discussed in Sec.\;\ref{SI: Sensitivity to Network Size and Training Period Length}.
%
Adding the neural closure to the low-fidelity model improved its matching with the high-fidelity data in nearly all cases. Its performance deteriorated with increasing the length of the time-sequences used to form the batches, and also with increasing the batch-size (the number of iterations per epoch is a dependent hyperparameter as mentioned earlier). This also led to an increase in training time. 
Depth of the networks affected the performance significantly, with shallower networks performing poorly than deeper networks as expected, however, the incremental gain in performance starts to taper off after certain depths (see Fig.\;\ref{fig: Exp_1_train_val_loss_all_network_size} and Sec.\;\ref{SI: Sensitivity to Network Size and Training Period Length}).
Using an exponentially decaying learning schedule over a constant learning rate tremendously improved learning performance and reduced the number of epochs needed. 
Further, training times slightly increased when using more delay times in the case of discrete-nDDE. In general, we found that the training time for discrete-nDDEs was similar to that for distributed-nDDEs.
Such behaviors by machine learning methods are difficult to anticipate in advance but they should be mentioned. 

Overall, in the experiments-1, we find that using memory-based neural closure models as we derived from the Mori-Zwanzig formulation is advantageous over just a Markovian closure. Using the new nDDEs as closure models helps maintain generalizability of the learned models for longer time-periods, and significantly reduces the longer-term prediction error of the ROM.  


\subsection{Experiments 2: Advecting Shock - Subgrid-Scale Processes}
\label{sec: exp2}
\begin{figure}[h!]
  \centering
  \subfloat[][]{\includegraphics[width=.4\textwidth]{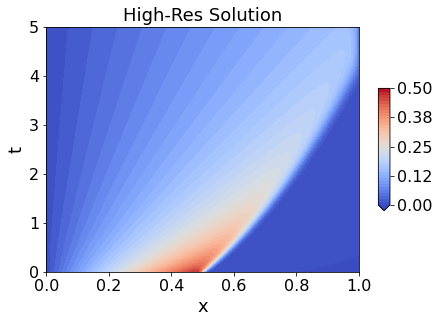}\label{fig: Exp2_high-res}} 
  \hspace{3em}%
  \subfloat[][]{\includegraphics[width=.4\textwidth]{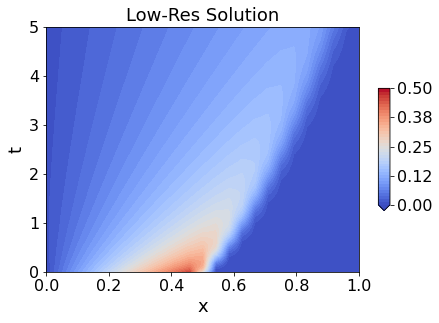}\label{fig: Exp2_low-res}}
  \\
  \subfloat[][]{\includegraphics[width=.4\textwidth]{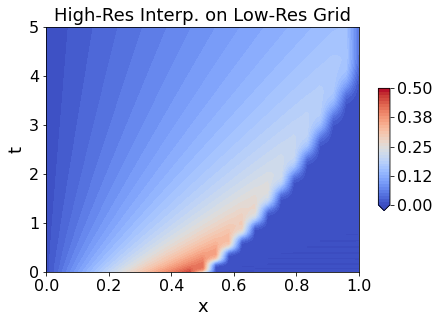}\label{fig: Exp2_high-res_interp_low-res_grid}}
  \hspace{3em}%
  \subfloat[][]{\includegraphics[width=.4\textwidth]{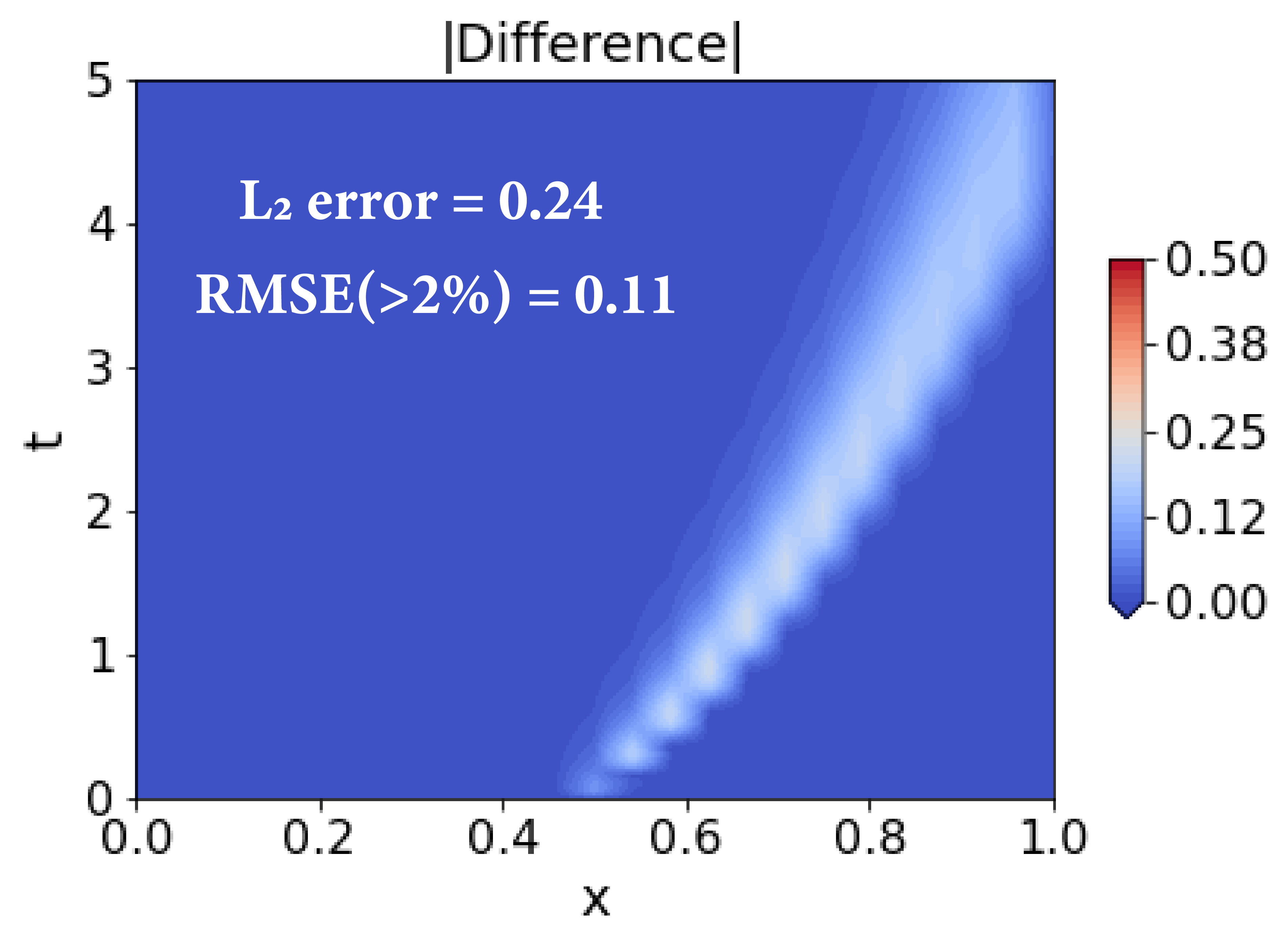}\label{fig: Exp2_high-res_low-res_diff}}
  \caption{Comparison of solutions of Burger's equation (Eq.\;\ref{eq: Advecting shock}) for different grid resolutions. \textit{(a):} Solution for a high-resolution grid with number of grid points, $N_x = 100$; \textit{(b):} Solution for a low-resolution grid with $N_x = 25$; \textit{(c):} High-resolution solution interpolated onto the low-resolution grid. \textit{(d):} Absolute difference between fields in panels (b) and (c). We also provide a pair of time-averaged errors, specifically: $L_2$ error; and RMSE considering only the grid points where the error is at least 2\% of the maximum velocity value, denoted by \textit{RMSE($>$2\%)}.
  }
  \label{fig: Exp2 Comparison of high and low-res solutions}
\end{figure}

In the second experiments, we again use the advecting shock problem governed by the Burger's equation (Eq.\;\ref{eq: Advecting shock}), but we now reduce the computational cost of the FOM by coarsening the spatial resolution, again leading to the need of a closure model (Sec.\;\ref{sec: Subgrid-Scale Processes}). For the high-fidelity/high-resolution solution, we employ a fine grid with the number of grid point in the $x$ direction $N_x = 100$, while for the low-fidelity/low-resolution solution, we employ a 4 times coarser grid with $N_x = 25$. 
A comparison of high- and low- resolution solutions solved using exactly the same numerical schemes and 
data stored at every
time-step of $\Delta t = 0.01$ is provided in Fig.\;\ref{fig: Exp2 Comparison of high and low-res solutions}. We observe that by decreasing the resolution, we introduce numerical diffusion and error in the location of the shock peak at later times. The goal of the neural closure models in these experiments is thus to augment the low-resolution model such that it matches the sub-sampled/interpolated high-resolution solution at the coarse (low-resolution) grid points.

For training our neural closure models for the low-resolution discretization with $N_x=25$, we use the same training regiment as in Experiments-1 (Sec.\;\ref{sec: exp1}), with architectures details presented in Table \ref{table: Exp1 and 2 architecture}. 
%
In order to exploit the fact that each grid point only affects its immediate neighbors over a single time-step, we use 1-D convolutional layers for these experiments.
%
For the nODE, we again employ a deeper architecture with more trainable parameters, and for the discrete-nDDE, six discrete delay values are again used.
%
The values of the other hyperparameters are in Sec.\;\ref{SI: hyperparameters}.
%
the validation period from $t = 1.25~\text{to}~2.5$, and the future prediction period from $t = 2.5~\text{to}~5.0$. 
We have chosen a prediction period of the combined length of training and validation periods.
For time-integration, we use the \textit{Vode} scheme \cite{brown1989vode} 
with adaptive time-stepping.
The true data are generated by interpolating the high-resolution solution onto the low-resolution grid $(\{\{u^{true}(x_k, T_i)\}_{k=1}^{N_x = 25}\}_{i=1}^M)$, as shown in Fig.\;\ref{fig: Exp2_high-res_interp_low-res_grid}, and we use the time-averaged $L_2$ error, $ \mathcal{L} = \frac{1}{M} \sum_{i=1}^M\left(\sqrt{\sum_{k=1}^{N_x=25} |u^{pred}(x_k, T_i) - u^{true}(x_k, T_i)|^2}\right) $, as the loss function.

The performance of the three neural closure models after 250 epochs (the stochastic gradient descent nearly converges, see Fig.\;\ref{fig: Exp_2_train_val_loss_all}) is evaluated by taking the absolute difference with the high-resolution solution interpolated onto the low-resolution grid (Fig.\;\ref{fig: Exp2_high-res_interp_low-res_grid}) 
spanning training, validation, and prediction periods.
We further benchmark our performance against the popular Smagorinsky model \cite{maulik2018explicit} used for subgrid-scale turbulence closure in large eddy simulation (LES). For the Burger's Eq.\;\ref{eq: Advecting shock}, it introduces a dynamic turbulent eddy viscosity ($\nu_e$) leading to, 
\begin{equation}
\label{eq: Advecting shock with Smagorinsky}
\frac{\partial u}{\partial t} + u\frac{\partial u}{\partial x} = \nu \frac{\partial^2 u}{\partial x^2} + \frac{\partial}{\partial x}\left( \nu_e \frac{\partial u}{\partial x}\right) \,,
\end{equation}
where $\nu_e = (C_s \Delta x)^2 \big|\frac{\partial u}{\partial x}\big|$ and $C_s$ is the Smagorinsky constant. 
Results are shown in Fig.\;\ref{fig: Exp2 Main results comparison between nODE, nDDE and nDistDDE}. The details of the architectures employed are in Table \ref{table: Exp1 and 2 architecture}. 
As shown by the error fields of the baseline (Fig.\;\ref{fig: Exp2_high-res_low-res_diff}) and closure models (Fig.\;\ref{fig: Exp2 Main results comparison between nODE, nDDE and nDistDDE}),
and by the corresponding pairs of averaged error numbers (see Figs.),
all closures improve the baseline.
%
However, the nODE and Smagorinsky closures only lead to a 55-60\% decrease in error, while the nDDE closures achieve a 80-90\% decrease. 
Despite the deeper architecture for the nODE, both the discrete-nDDE and distributed-nDDE (with smaller architectures) again achieve smaller errors, for the whole period of $t=0~\text{to}~5.0$. This means that they have lower numerical diffusion, thus capturing the targeted subgrid-scale process. 
%
As opposed to the findings of experiments-1 (Fig.\;\ref{fig: Exp1_Comparison_Coeff_Plots}), in the present experiments-2, the discrete-nDDE performs slightly better than the distributed-nDDE in the prediction period.

\begin{figure}
  \centering
  \subfloat[][Smagorinsky LES model]{\includegraphics[width=0.7\textwidth]{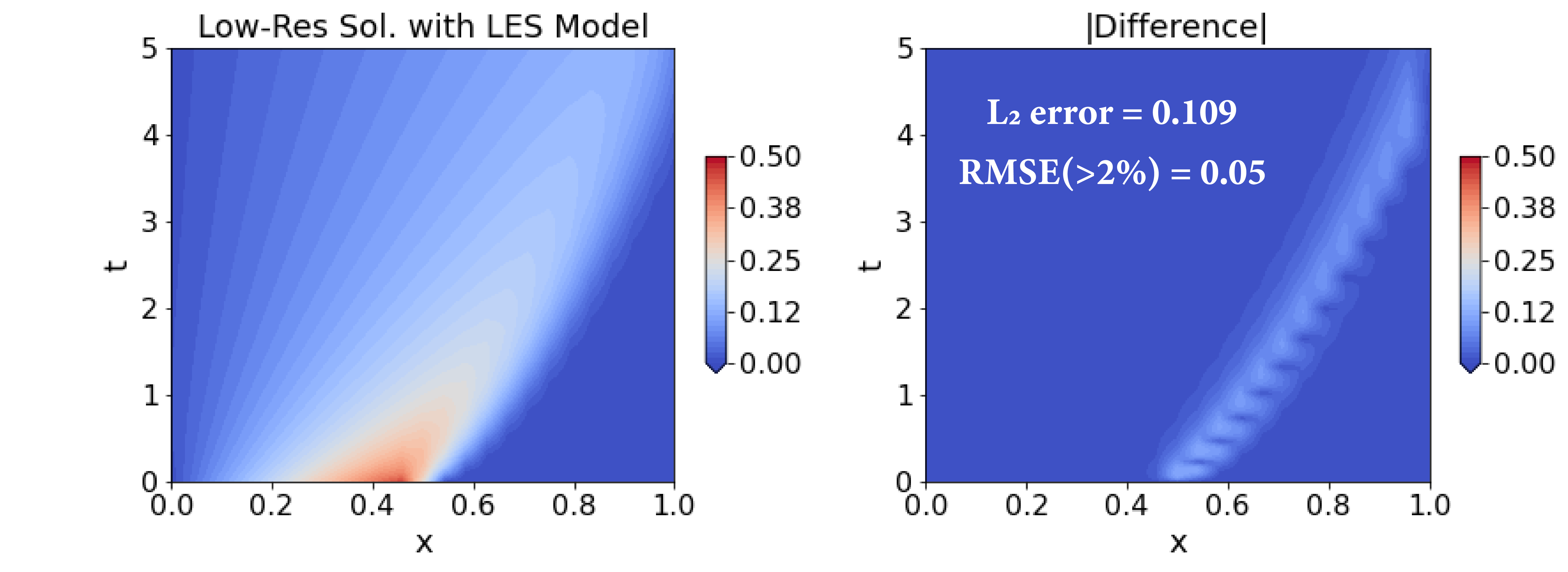}\label{fig: Exp2_LES_result_modified}} 
  \\
  \subfloat[][Neural closure model with no-delays (nODE)]{\includegraphics[width=0.7\textwidth]{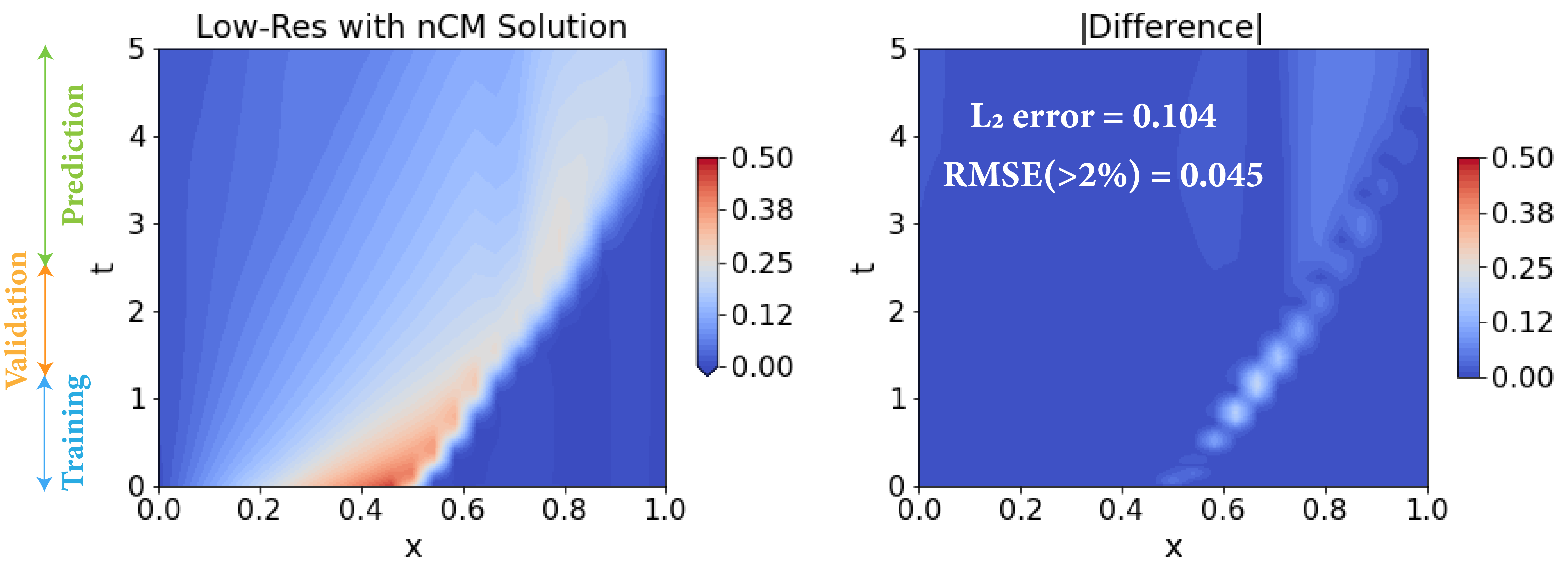}\label{fig: Exp2_nODE_result_modified}} 
  \\
  \subfloat[][Neural closure model with discrete-delays (Discrete-nDDE)]{\includegraphics[width=0.7\textwidth]{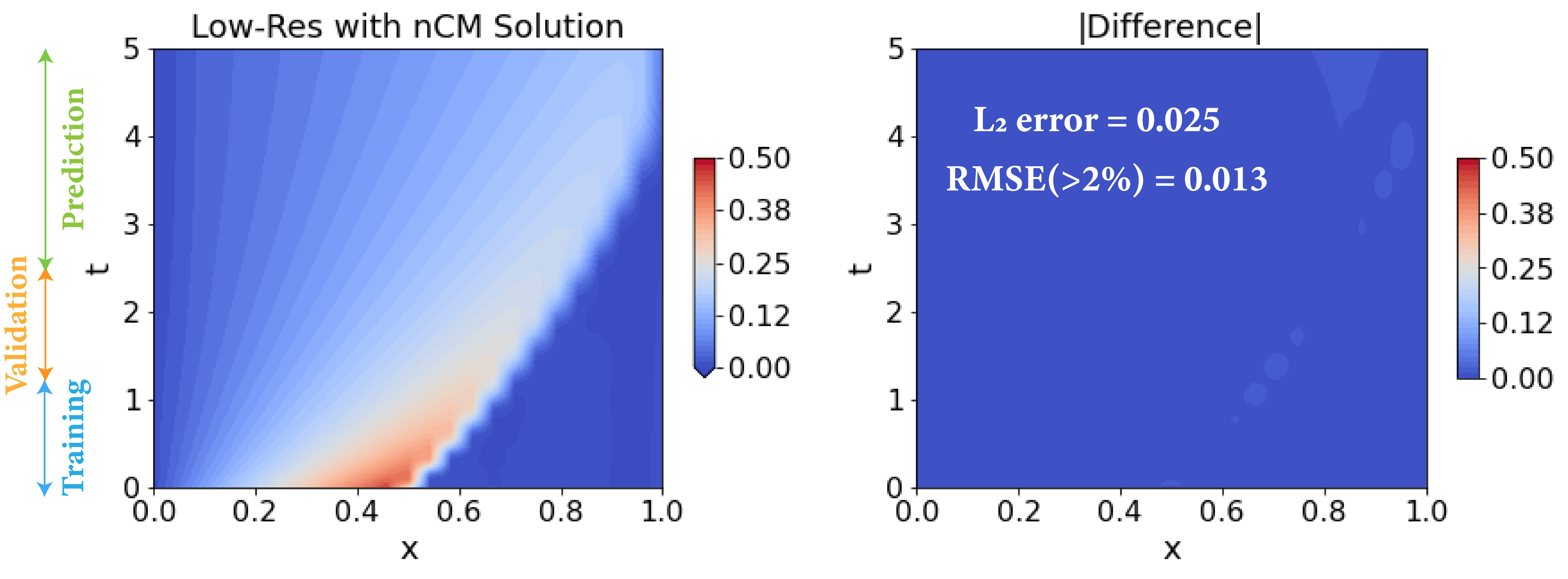}\label{fig: Exp2_nDDE_result}}  \\
  \subfloat[][Neural closure model with distributed-delays (Distributed-nDDE)]{\includegraphics[width=0.7\textwidth]{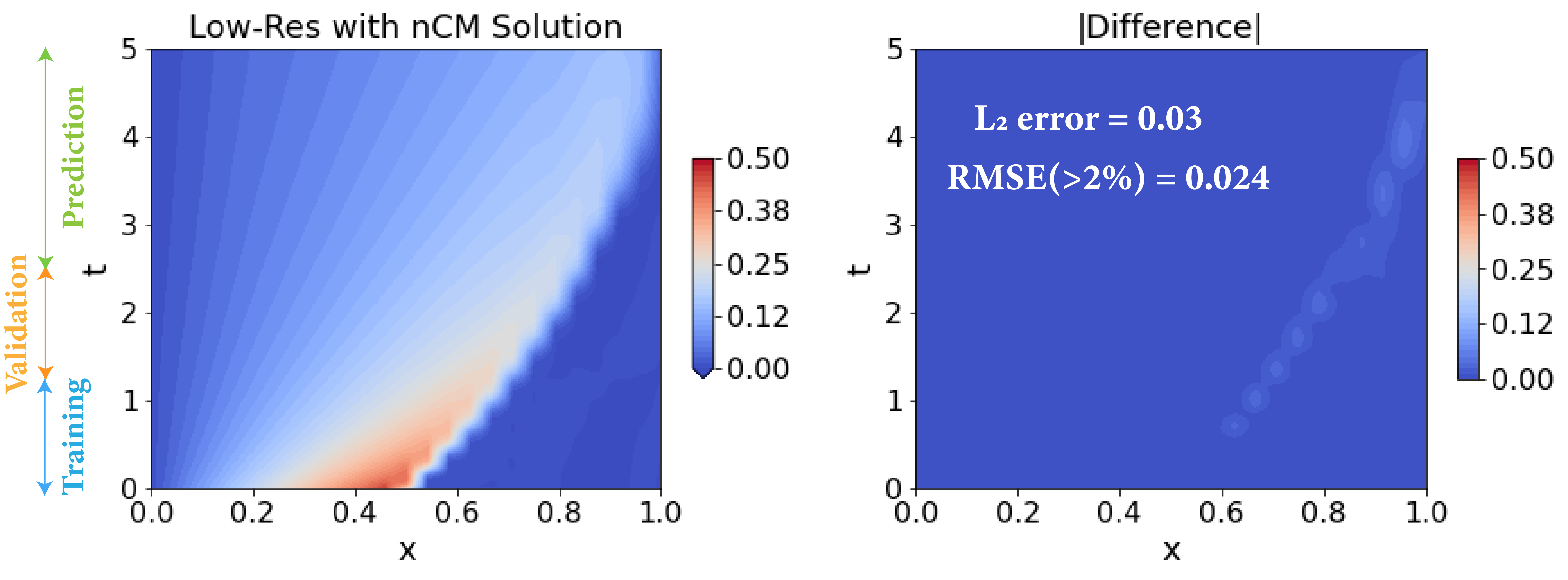}\label{fig: Exp2_nDistDDE_result}}
  \caption{Solutions of the Burger's PDE on the low-resolution grid with different closure models (\textit{left-column}), and their absolute differences (\textit{right-column}) with the high-resolution solution interpolated onto the low-resolution grid (Fig.\;\ref{fig: Exp2_high-res_interp_low-res_grid}). %
  For the trained neural closure models, 
  the training period is from $t=0~\text{to}~1.25$, the validation period from $t=1.25~\text{to}~2.5$, and the prediction period from $t=2.5~\text{to}~5.0$.
  For each closure, we also provide the pair of time-averaged errors (see Fig.\;\ref{fig: Exp2 Comparison of high and low-res solutions} for description).
  \textit{(a):} Smagorinsky LES model with $C_s = 1.0$;
  \textit{(b)}, \textit{(c)}, \textit{(d)}: different neural closure models. These results correspond to the architectures detailed in Table \ref{table: Exp1 and 2 architecture}.
  }
  \label{fig: Exp2 Main results comparison between nODE, nDDE and nDistDDE}
\end{figure}

\begin{figure}[!t]
  \centering
  \subfloat[][Experiments - 2]{\includegraphics[width=.40\textwidth]{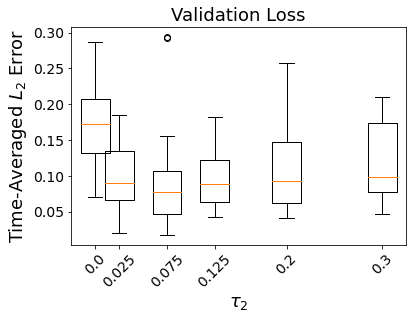}\label{fig: Exp2_optimal_delay_barplot}}\qquad
  \subfloat[][Experiments - 3a]{\includegraphics[width=.40\textwidth]{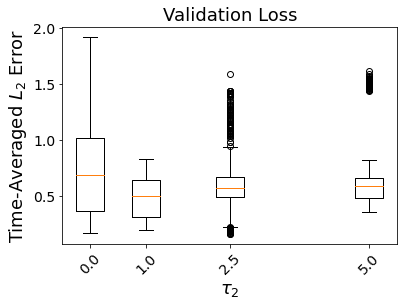}\label{fig: Exp3_optimal_delay_barplot}}
  \caption{
  Variation of distributed-nDDE closure validation loss (time-averaged $L_2$ error) averaged over the last 50 training epoch for Experiments-2 \& 3a. All the experiments have $\tau_1 = 0$, and different $\tau_2$ (horizontal-axis). Note that $\tau_2=0$ corresponds to the nODE closure. 
  We use boxplots to provide statistical summaries for multiple training repeats done for each experiment. The box and its whiskers provide a five number summary: minimum, first quartile (Q1), median (orange solid line), third quartile (Q3), and maximum, along with outliers (black circles) if any.  
  }
  \label{fig: Prediction loss for nDDE-Dist different delays comparison}
\end{figure}

We now study the effect of changing the amount of past information incorporated in the closure model on the time-averaged $L_2$ error. For this, we fix $\tau_1 = 0$ for the distributed-nDDE closure, and vary the values of only $\tau_2$, keeping the architecture the same (Table \ref{table: Exp1 and 2 architecture}). 
%
First, for the time-averaged training loss (not shown), we found no discernible trend and differences were mostly due the stochastic gradient descent.
Second, for the validation period, Fig.\;\ref{fig: Exp2_optimal_delay_barplot} shows the statistical summary of the validation losses (time-averaged $L_2$ error) 
between the last epochs $200$ to $250$, for different delay-period lengths. In order to ensure statistical soundness of the results, $10-15$ repeats of the training were done, and the results aggregated for each delay-period length.
%
%
Results indicate that the validation loss first decreases and then increases as we incorporate more-and-more past information, starting from a very small delay period. For a specified architecture, neither too little nor too much past information is helpful: there is likely an optimal amount of information to incorporate.
The initial improvement in the performance of the closure models as the delay period is increased is due to the increase in information content about 
the recent past. However, a particular network architecture of finite size will have a limit on capturing the increasing information content effectively due to its limited expressive power, thus leading to a decrease in performance when too much information is provided. %
An estimate of the range of delay period lengths to consider can be obtained from properties of the given dynamical system such as the main time-scales, e.g.\ advection and diffusion times-scales in the present system, and main decorrelation times of state variables. Overall, from Fig.\;\ref{fig: Exp2_optimal_delay_barplot}, we can notice the optimal delay period length to be around $0.075$. Similar trends between performance of neural closure models and delay period lengths were also found in Experiments-1 (not shown).
%
Some published studies attempt to derive analytical expressions for the optimal memory length, making many approximations in the process \cite{li2019mori, pan2018data}. 
%
A final option is to learn the delay amount as a part of the training process itself, however, this requires modified adjoint equations.

We conducted again a series of experiments-2 to assess the sensitivity to the various other hyperparameters, and found similar trends (not shown here) as in the experiments-1. 
Finally, we noticed that using the \textit{dopri5} \cite{hairer1993solving} time-integration scheme severely impaired the learning ability in the experiments-2.

Overall, these results demonstrate the superiority of using our new memory-based closure models in capturing subgrid-scale processes.



\subsection{Experiments 3a: 0-D Marine biological Models}
\label{sec:biogeochemical 0-D models}
For our third experiments, we use neural closure models to incorporate the effects of missing processes and state variables in lower-complexity biological models, thus targeting the third class of 
closure modeling (Sec.\;\ref{sec: Complexity of Models}). Marine biological models are based on ODEs that describe the food-web interactions in the ecosystem. They can vary greatly in terms of complexity \cite{fennel2014introduction}.
The marine biological models used in our experiments are adapted from Newberger et. al., 2003 (\cite{newberger2003analysis}).
They were used to simulate the ecosystem in the Oregon coastal upwelling zone.
They provide hierarchical embedded models compatible with each other. 
We employ the three-component NPZ model (nutrients ($N$), phytoplankton ($P$), and zooplankton ($Z$)), and the five-component NNPZD model (ammonia ($NH_4$), nitrate ($NO_3$), $P$, $Z$, and detritus ($D$)) 
in a zero-dimensional setting (0-D; only temporal variation).
The low complexity NPZ model is given by,
\begin{equation}
\label{eq:NPZ model}
\begin{split}
\frac{dN}{dt} &= -G\frac{PN}{N+K_u}+\Xi P + \Gamma Z + R_m \gamma Z(1-\exp^{-\Lambda P})  \\
\frac{dP}{dt} &= G\frac{PN}{N+K_u} - \Xi P  - R_m Z(1-\exp^{-\Lambda P})  \\
\frac{dZ}{dt} &=   R_m (1-\gamma) Z(1-\exp^{-\Lambda P}) - \Gamma Z \\
\text{with} \quad N(0) &= T_{bio}, \quad P(0) = 0, \quad \text{and} \quad Z(0) = 0 \,,
\end{split} 
\end{equation}
%
with $G$ representing the optical model,
\begin{equation}
\label{eq: optical model}
    G = V_m \frac{\alpha I}{(V_m^2 + \alpha^2 I^2)^{1/2}}, \quad \text{and} \quad I(z) = I_0 \exp (k_w z) \, ,
\end{equation}
where $z$ is depth and $I(z)$ models the availability of sunlight for photo-chemical reactions. The parameters in Eqs.\;\ref{eq:NPZ model} and \ref{eq: optical model} are: $k_w$, light attenuation by sea water; $\alpha$, initial slope of the $P-I$ curve; $I_0$, surface photosynthetically available radiation; $V_m$, phytoplankton maximum uptake rate; $K_u$, half-saturation for phytoplankton uptake of nutrients; $\Xi$, phytoplankton specific mortality rate; $R_m$, zooplankton maximum grazing rate; $\Lambda$, Ivlev constant; $\gamma$, fraction of zooplankton grazing egested; $\Gamma$, zooplankton specific excretion/mortality rate; and $T_{bio}$, total biomass concentration. 
In the NPZ model (Eq.\;\ref{eq:NPZ model}), the nutrient uptake by phytoplankton is governed by a Michaelis-Menten formulation, which amounts to a linear uptake relationship at low nutrient concentrations that saturates to a constant at high concentrations.
%
The grazing of phytoplankton by zooplankton follows a similar behavior: their growth rate becomes independent of $P$ in case of abundance, but proportional to available $P$ when resources are scarce, hence zooplankton grazing is modeled by an Ivlev function. The death rates of both $P$ and $Z$ are linear, and a portion of zooplankton grazing in the form of excretion goes directly to nutrients.

In the higher complexity NNPZD model, the nutrients are divided into ammonia and nitrates, which are the two most important forms of nitrogen in the ocean.  With the intermediate of decomposed organic matter, detritus, the NNPZD model captures new processes such as: phytoplankton cells preferentially taking up ammonia over nitrate because the presence of ammonia inhibits the activity of the enzyme nitrate reductase essential for the uptake kinetics; the pool of ammonium coming from remineralization of detritus; and part of this ammonium pool getting oxidized to become a source of nitrate called the process of nitrification, etc. 
Overall, the NNPZD model is given by,
\begin{equation}
\label{eq:NNPZD Model}
\begin{split}
\frac{d NO_3}{dt} =& ~\Omega NH_4 - G\left[\frac{NO_3}{NO_3+K_u}\exp^{-\Psi NH_4}\right]P   \\
\frac{d NH_4}{dt} =& ~-\Omega NH_4 + \Phi D +\Gamma Z  - G\left[\frac{NH_4}{NH_4+K_u}\right]P   \\
\frac{dP}{dt} =& ~G\left[\frac{NO_3}{NO_3+K_u}\exp^{-\Psi NH_4} + \frac{NH_4}{NH_4+K_u}\right]P - \Xi P  - R_m Z(1-\exp^{-\Lambda P})  \\
\frac{dZ}{dt} =&  ~ R_m (1-\gamma) Z(1-\exp^{-\Lambda P}) - \Gamma Z   \\
\frac{dD}{dt} =&~ R_m \gamma Z(1-\exp^{-\Lambda P}) + \Xi P - \Phi D \\
\text{with} \quad &NO_3(0) = T_{bio}/2, \quad NH_4(0) = T_{bio}/2, \quad P(0) = 0, \quad Z(0) = 0, \;\; \text{and} \;\; D(0) = 0 \,,
\end{split}
\end{equation} 
where the new parameters are: $\Psi$, $NH_4$ inhibition parameter; $\Phi$, detritus decomposition rate; and $\Omega$, $NH_4$ oxidation rate.

\begin{figure}[h!]
  \centering
  \subfloat[][NPZ Model]{\includegraphics[width=.425\textwidth]{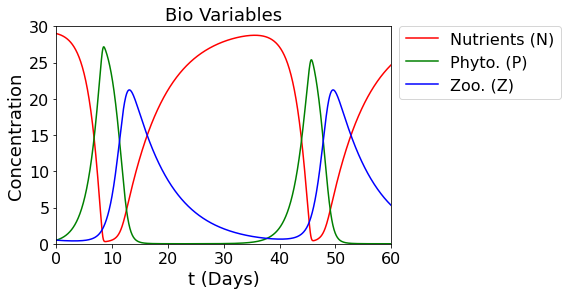}\label{fig: Exp3 npz model}} 
  \quad~~
  \subfloat[][NNPZD Model] {\includegraphics[width=.425\textwidth]{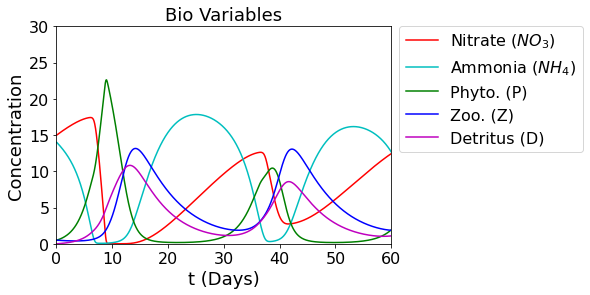}\label{fig: Exp3 nnpzd model}} \\
  \subfloat[][Comparison of NPZ and NNPZD Models]{\includegraphics[width=.5\textwidth]{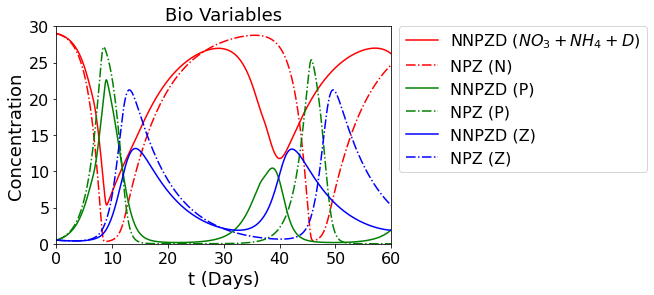}\label{fig: Exp3 npz nnpzd comparison}}
  \caption{
  Solutions of the marine biological models used in Experiments-3a (concentrations vs.~time in \textit{days}). Parameter values used are (adopted from \cite{newberger2003analysis}): $k_w = 0.067~m^{-1}$, $\alpha = 0.025 ~(W~m^{-2}~d)^{-1}$, $V_m = 1.5~ d^{-1}$, $I_0 = 158.075 ~W~m^{-2}$, $K_u = 1~ mmol~N~m^{-3}$, $\Psi = 1.46~(mmol~N~m^{-3})^{-1}$, $\Xi = 0.1 ~d^{-1}$, $R_m = 1.52 ~ d^{-1}$, $\Lambda = 0.06~(mmol~N~m^{-3})^{-1}$, $\gamma = 0.3$, $\Gamma = 0.145~d^{-1}$, $\Phi = 0.175~d^{-1}$, $\Omega = 0.041 ~d^{-1}$, $z = -25~m$, and $T_{bio} = 30~ mmol~N~m^{-3}$. \textit{(a):} Nutrient-Phytoplankton-Zooplankton (NPZ) model (Eq.\;\ref{eq:NPZ model}); \textit{(b):} Nitrate-Ammonia-Phytoplankton-Zooplankton-Detritus (NNPZD) model (Eq.\;\ref{eq:NNPZD Model}); \textit{(c):} Comparison between $NO_3+NH_4+D$, $P$, and $Z$ from the NNPZD model (\textit{solid}) with $N$, $P$ and $Z$ from the NPZ model (\textit{dashed-dot}).}
  \label{fig: Bio model response}
\end{figure}

Solutions of the above two models are presented in Fig.\;\ref{fig: Bio model response}. Different values of the parameters and initial conditions set these models in different dynamical regimes.
From the responses in time, the present solutions in experiments-3a are in a stable nonlinear limit-cycle regime. The $N$ class in the NPZ model is a broader class encompassing $NO_3$, $NH_4$, and $D$ from the NNPZD model. Since the NNPZD model resolves many more processes, the concentrations of $NO_3+NH_4+D$, $P$, and $Z$ differ significantly from the $N$, $P$, and $Z$ of the NPZ model. 
The goal of the neural closure models in these experiments is thus to augment the low-complexity NPZ model such that it matches the aggregated states of the high-complexity NNPZD model.

For training our neural closure models for the NPZ model, we use the same training regiment as in Experiments-1 \& 2 (Secs.\;\ref{sec: exp1} \& \ref{sec: exp2}), with architectures details presented in Table \ref{table: Exp3 and 4 architecture}. 
For the nODE, we again employ a bigger  architecture, and for the discrete-nDDE, six discrete delay values are again used. 
%
The values of other hyperparameters are given in Sec.\;\ref{SI: hyperparameters}.
%
The training period ranges from $t = 0~\text{to}~30~days$, validation period from $t = 30~\text{to}~60~days$; and the prediction period from $t = 60~\text{to}~330~days$. We have chosen a prediction period nine times longer than the training period.
%
%
%
For biological ODE models, there exists invariant knowledge about the system, such as biological state variables cannot be negative, and the sum of all the states remains constant with time (this can be verified by summing the ODEs of NPZ or NNPZD models). 
%
We enforce the constraints as follows.
The positivity is enforced as a penalization term in the loss function. The constant total biomass constraint is embedded in the architectures of neural closures by introducing a new custom layer named, \textit{BioConstrainLayer}. This layer is applied at the end, and expects an input of size 1. The output of this layer is formed by splitting the input into three with the proportions, $\beta$, $-1$, and $1-\beta$; where $\beta$ is a trainable parameter. This ensures that summing the right hand side of the augmented NPZ model does not leave any new residual due to the neural closure terms.
The stiffness of such biological ODE models also poses a challenge in maintaining these desired properties \cite{burchard2005application}. The flexibility of our framework however allows the use of appropriate time-stepping schemes, such as A-stable implicit schemes, etc., to overcome stiffness.
%
The true data are generated by aggregating the variables of the NNPZD model $(N \equiv NO_3+NH_4+D,~P,~\text{and}~Z;~\{ \{ B^{true}(T_i)\}_{B\in\{N, P, Z\}}\}_{i=1}^M)$. Finally, we use the \textit{dopri5} \cite{hairer1993solving} scheme with adaptive time-stepping and simulation data were stored at every  $\Delta t = 0.05$~days for all our time-integration requirements, along with a $L_2$ error loss function, $ \mathcal{L} = \allowdisplaybreaks  \frac{1}{M} \sum_{i=1}^M\left(\sqrt{\sum_{B\in\{N, P, Z\}} |B^{pred}(T_i) - B^{true}(T_i)|^2}\right) $.

\begin{figure}
  \centering
  \subfloat[][Neural closure model with no-delays (nODE)]{\includegraphics[width=1\textwidth]{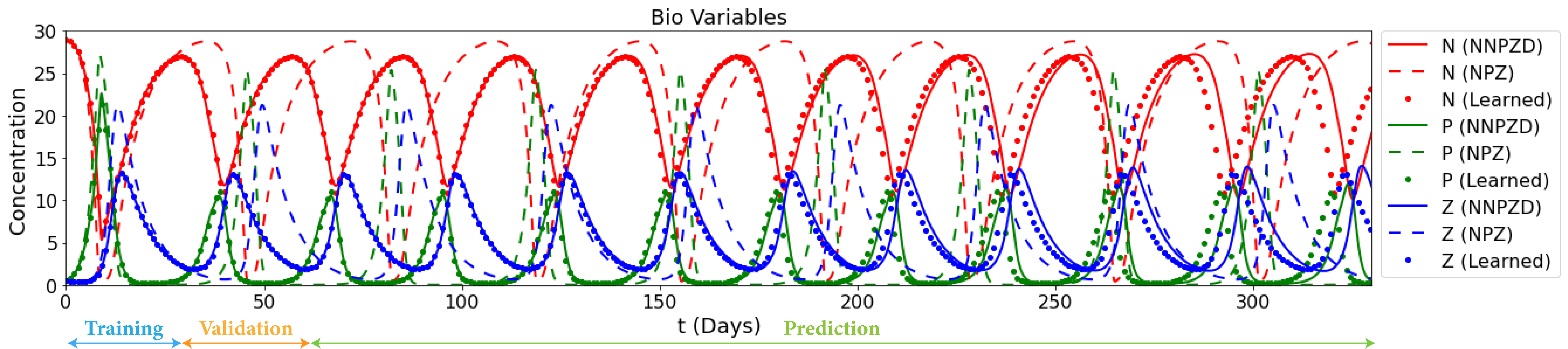}\label{fig: Exp3_nODE_result_modified}} 
  \\
  \subfloat[][Neural closure model with discrete-delays (Discrete-nDDE)]{\includegraphics[width=1\textwidth]{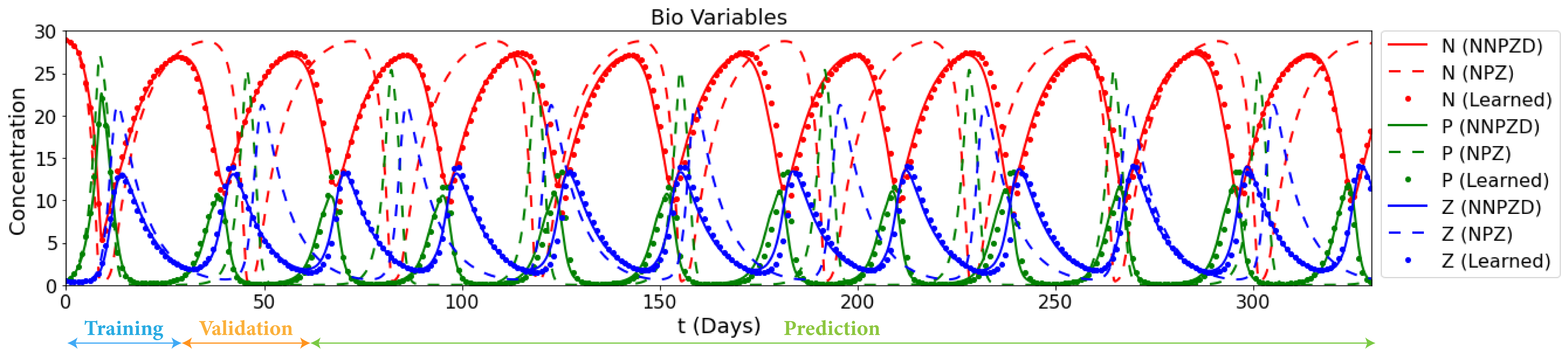}\label{fig: Exp3_nDDE_result_modified}} 
  \\
  \subfloat[][Neural closure model with distributed-delays (Distributed-nDDE)]{\includegraphics[width=1\textwidth]{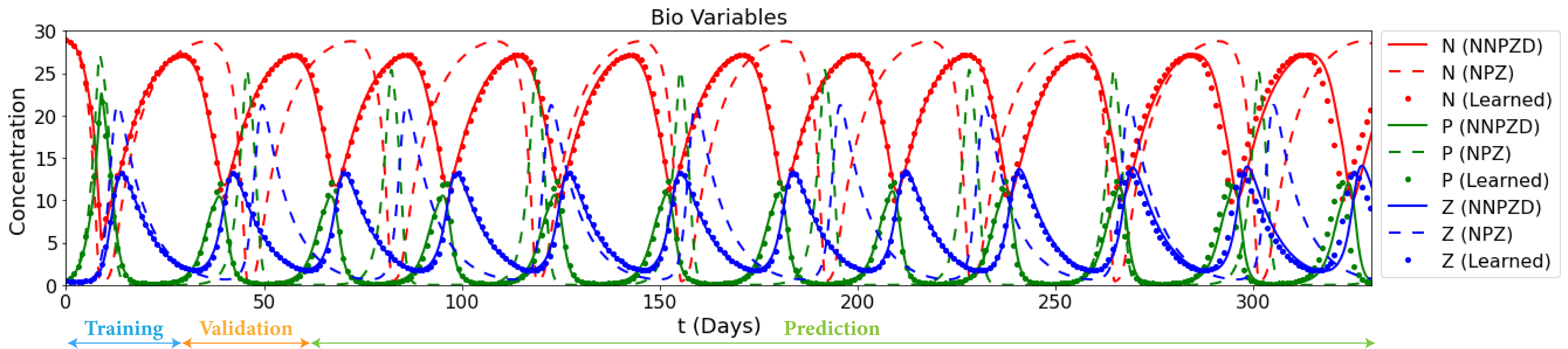}\label{fig: Exp3_nDistDDE_result_modified}} 
  \\
  \subfloat[][Performance comparison of different neural closure models]{\includegraphics[width=1\textwidth]{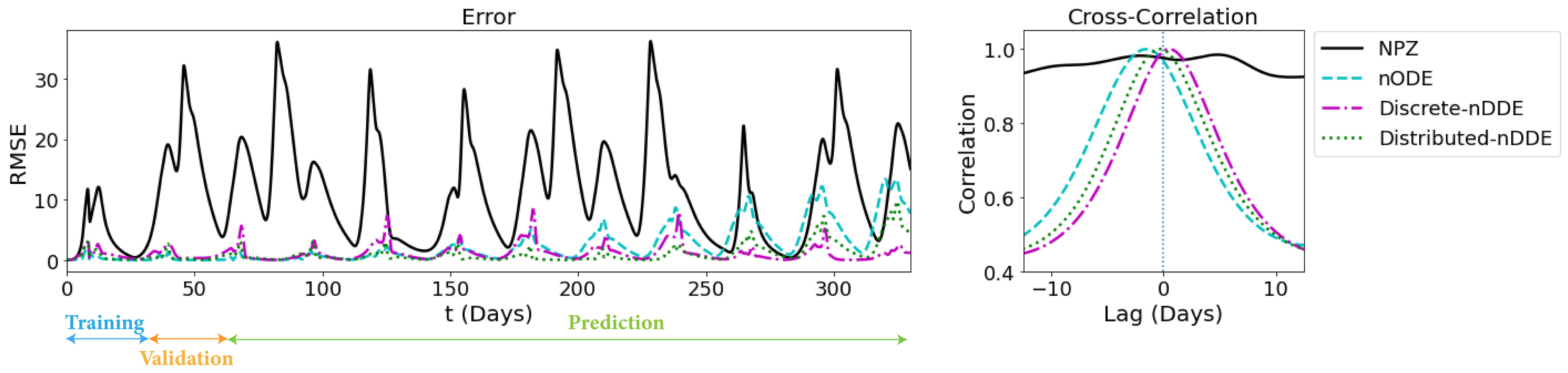}\label{fig: Exp3_RMSE_xcorr_modified}} 
  \\
  \caption{
  Comparison of the biological variables from the learned NPZ model augmented with the three neural closure models (\textit{dashed}), aggregated variables from the NNPZD model (ground truth; \textit{solid}), and variables from the NPZ model (\textit{dashed-dot}) at the end of training. 
  For each neural closure, the training period is from $t=0~\text{to}~30$ days, the validation period is from $t=30~\text{to}~60$ days, while prediction period is from $t=60~\text{to}~330$ days.
  \textit{(a)}, \textit{(b)}, \textit{(c):} different neural closure models; 
  \textit{(d):} the \textit{left} plot shows the evolution of root-mean-squared-error (RMSE), and the \textit{right} plot shows the average cross-correlation (only for the prediction period) w.r.t. the ground truth. These results correspond to the architectures detailed in Table \ref{table: Exp3 and 4 architecture}.
  }
  \label{fig: Exp3_Results_Combined}
\end{figure}

The performance of the three neural closure models augmenting the NPZ is evaluated after 350 epochs of training (the stochastic gradient descent nearly converges, as evident from the Fig.\;\ref{fig: Exp_3_train_val_loss_all}) by comparison with the aggregated biology variables from the high-complexity NNPZD model (Eq.\;\ref{eq:NNPZD Model}) 
spanning training, validation, and prediction periods.
Results are presented in Figure \ref{fig: Exp3_Results_Combined}. The details of the architectures employed are in Table \ref{table: Exp3 and 4 architecture}. 
%
When compared with the aggregated NNPZD variables (true variables), we find again that despite the bigger architecture of the nODE, it starts to develop significant errors around $t = 180~days$ and quickly gets out-of-phase
thereafter (Fig.\;\ref{fig: Exp3_nODE_result_modified}). 
The discrete-nDDE and distributed-nDDE, both with smaller architectures, are however able to match the true variables for nearly the whole period of $t=0~\text{to}~330 ~days$ (Fig.\;\ref{fig: Exp3_nDDE_result_modified}), with only distributed-nDDE starting to getting out-of-phase after $t = 270~days$ (Fig.\;\ref{fig: Exp3_nDistDDE_result_modified}) at the end of the long prediction period. These results are corroborated by the time evolution of the RMSE and average cross-correlation for the three variables over the prediction period (Fig.\;\ref{fig: Exp3_RMSE_xcorr_modified}).
From the progression of the time-averaged $L_2$ loss (here, the error between the variables from the closure-model-augmented NPZ system, and the true variables), the nODE performs either equally well or better than both discrete-nDDE and distributed-nDDE during training and validations periods (Fig.\;\ref{fig: Exp_3_train_val_loss_all}), however, it is not able to maintain long-term accuracy. We also notice very large spikes in the first half of the training regime, which are due to weights of the neural-networks taking values that lead to negative biology variables. As training progresses, we however don't observe this behavior anymore because the trainable weights starts to converge towards biologically feasible regimes.
%
%
In conclusion, using a memory based closure for a low-complexity model can efficiently help emulate the high-complexity model dynamics.

As for experiments-2, we conducted a series of experiments-3a, to study the effect of changing the amount of past information incorporated in the neural closure models. 
In Fig.\;\ref{fig: Exp3_optimal_delay_barplot}, we show the variation of the  average validation loss (time-averaged $L_2$ error) between the last epochs $300$ to $350$, for different delay-period lengths ($\tau_1 = 0$, and $\tau_2$ varying in case of distributed-nDDE). In order to ensure statistical soundness of the results, $10-12$ repeats of the training were done, and the results were aggregated for each delay-period length (excluding the runs which diverged). We again find an optimal memory length for a specified architecture, however, with more and more runs failing to converge for longer delay period lengths.
For the present system, estimates of delay period lengths to consider can be obtained from the time-scales of biological behaviors and adjustments, and from the decorrelation times of the biological state variables. Taking into account the limited effectiveness of a network architecture of finite size for capturing increasing information content, from Fig.\;\ref{fig: Exp3_optimal_delay_barplot}, we find an optimal delay period length to be around $1~day$.
We also conducted a series of experiments-3a to study the sensitivity to the various hyperparameters, and found similar trends (not shown here) as in experiments-1\;\&\;2.
%
For good performance, we further found that using a small enough time-step was critical as well as limiting the number of internal steps in the \textit{dopri5} \cite{hairer1993solving} time-integration scheme,
%
while penalizing negative values in the loss function did not make much of a difference. Whenever multiple terms are present in the loss function enforcing different inherent properties of the system, they should be normalized (e.g.\;using non-dimensional variables) and given appropriate relative weights.  
%

In general, the ecosystem ODEs are coupled with regional or global ocean modeling systems, leading to advection-diffusion-reaction PDEs \cite{besiktepe_et_al_JMS2003}. If highly complex ecosystem models are employed, a very large number of PDE state variables need to be solved for, rendering the computations very expensive. A large number of unknown parameter values as well as uncertain initial conditions then also need to be estimated, requiring specific methods \cite[e.g.][]{lermusiaux_et_al_Oceanog2011}.
The available biogeochemical observations are not always sufficient for calibrating these many unknown parameters and for estimating initial conditions of high-complexity models. If the corresponding errors are large, this can lead to integrating models in the wrong dynamical regimes \cite[e.g.][]{ueckermann_lermusiaux_OD2010}.
Finally,
in some applications, one is only interested in the aggregated state variables as the output, but cannot use low-complexity models because their dynamics are too inaccurate for the goals of the applications. 
Using neural closure models as shown here, one can increase the accuracy of the low-complexity models to match the response of high-complexity models (possibly up to models such as ERSEM \cite{baretta1995european}) without adding the computational burden of modeling all the intermediate biological states and processes, while reducing the effects of other uncertainties listed above. Results of our neural closures in 1-D PDEs is showcased next.

\subsection{Experiments 3b: 1-D Marine Biogeochemical Models}
\label{sec:biogeochemical 1-D models}

\begin{figure}
  \centering
  \subfloat[][Aggregated NNPZD model states (ground truth)]{\includegraphics[width=0.725\textwidth]{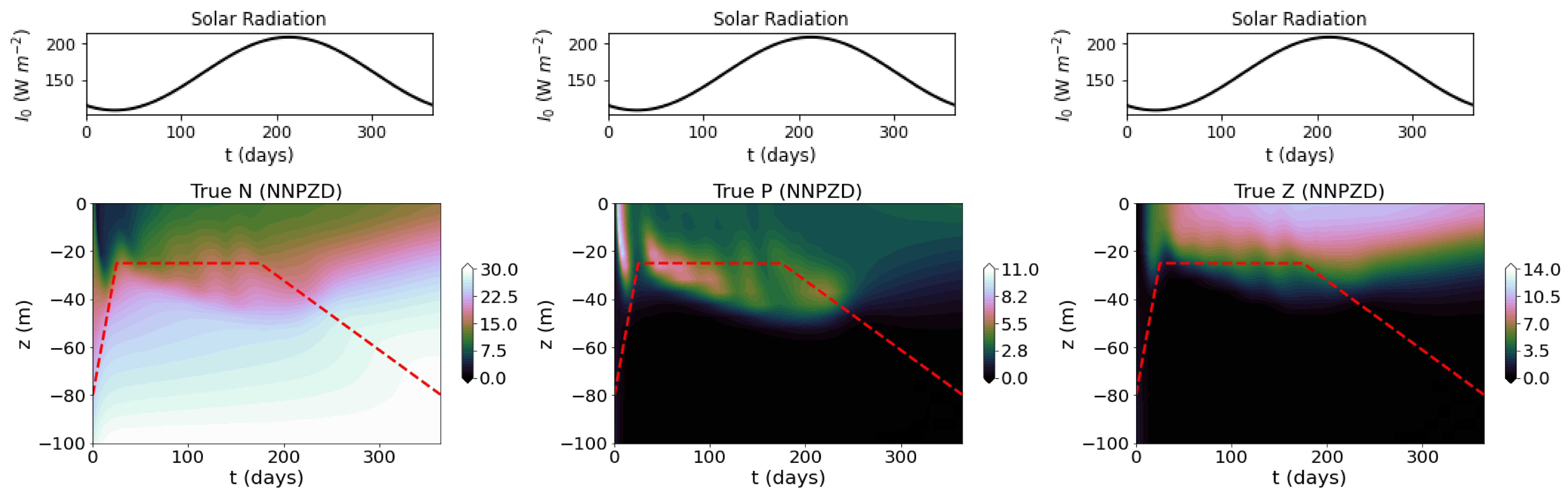}\label{fig: Exp4_npz_agg_sol_modified}} 
  \\
  \subfloat[][NPZ model (without neural closure model)]{\includegraphics[width=0.725\textwidth]{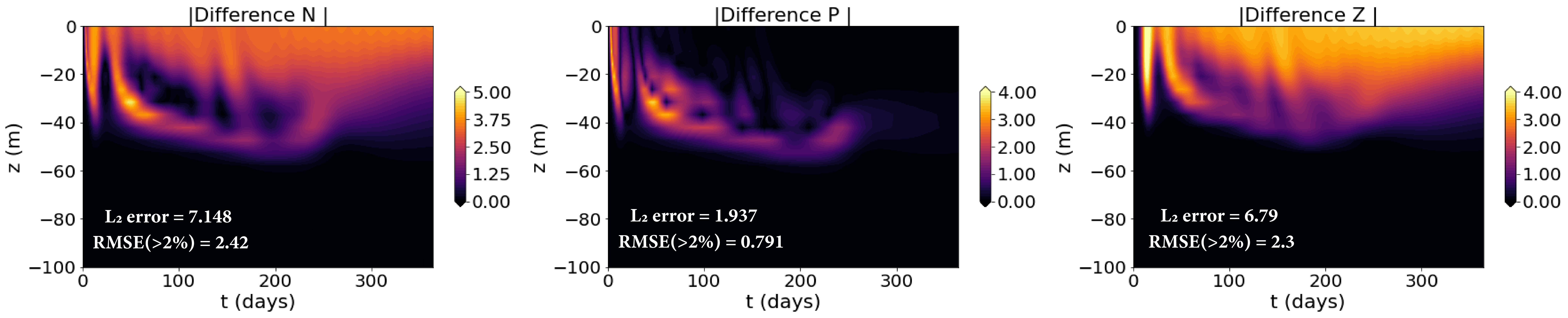}\label{fig: Exp4_npz_nnpzd_diff_modified}} 
  \\
  \subfloat[][NPZ model augmented with no-delay neural closure (nODE)]{\includegraphics[width=0.725\textwidth]{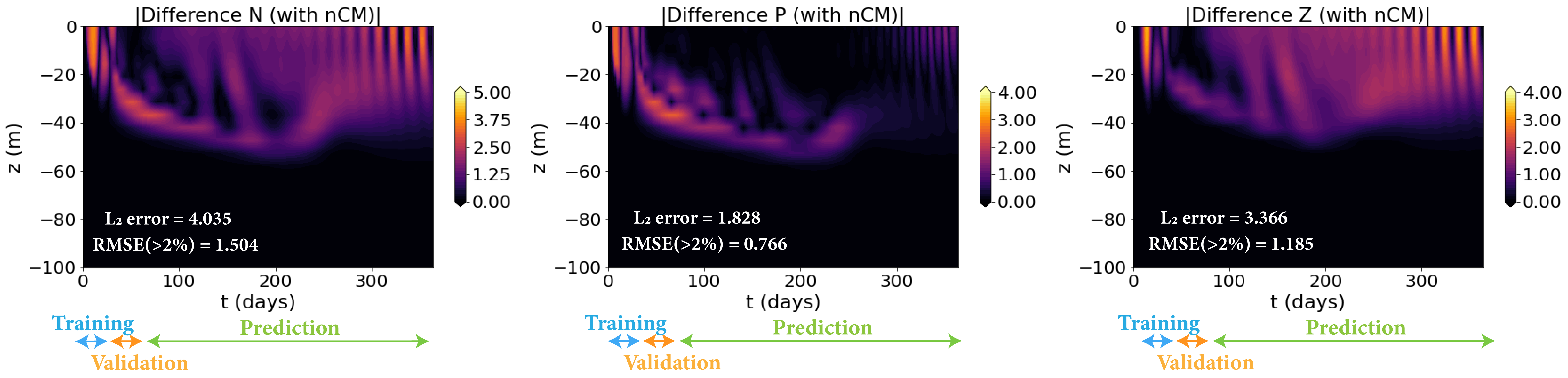}\label{fig: Exp4_nODE_result_modified}}\\
  \subfloat[][NPZ model augmented with discrete-delay neural closure (Discrete-nDDE)]{\includegraphics[width=0.725\textwidth]{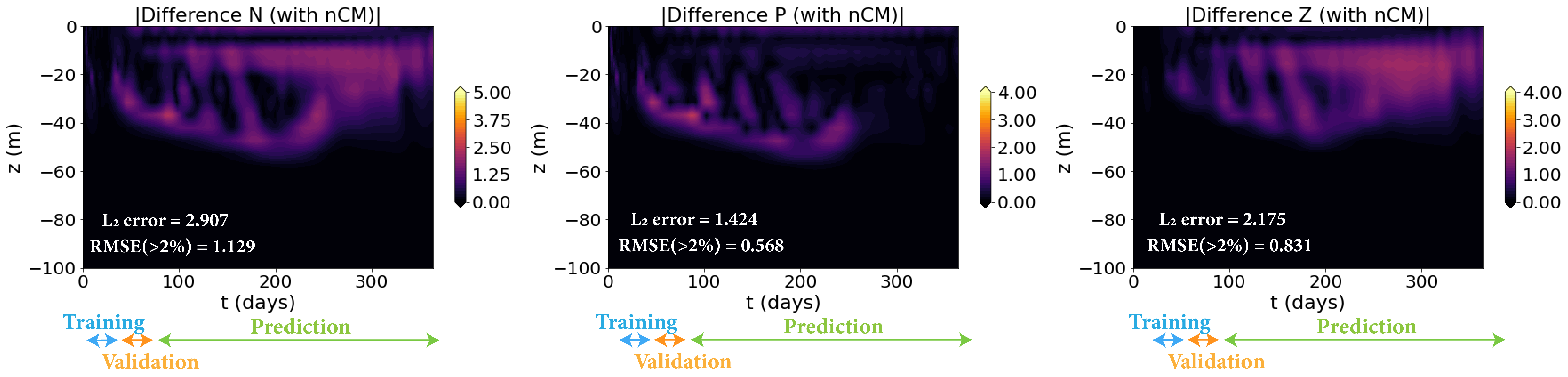}\label{fig: Exp4_nDDE_result_modified}}  \\
  \subfloat[][NPZ model augmented with distributed-delay neural closure (Distributed-nDDE)]{\includegraphics[width=0.725\textwidth]{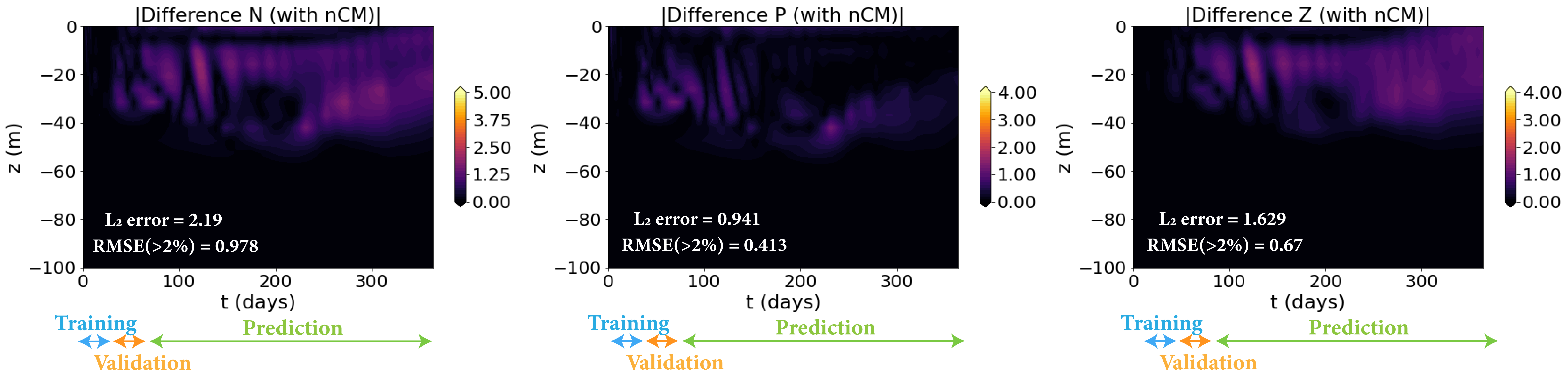}\label{fig: Exp4_nDistDDE_result_modified}}
  \caption{
  Comparison of the 1-D physical-biogeochemical PDE models used in Experiments-3b with and without closure models. %
  Along with the parameter values mentioned in Figure~\ref{fig: Bio model response}, we consider: a sinusoidal variation in $I_o(t)$; linear vertical variation in total biomass $T_{bio}(z)$ from $10~mmol~N~m^{-3}$ at the surface to $30~mmol~N~m^{-3}$ at $z = 100~m$; $K_{z_b} = 0.0864~(m^2/day)$; $K_{z_0} = 8.64~(m^2/day)$; $\gamma = 0.1~m^{-1}$; and $D = -100~m$, all adapted from \cite{newberger2003analysis, eknes2002ensemble}.
  For the neural closure models, the training period is from $t=0~\text{to}~30$ days, the validation period from $t=30~\text{to}~60$ days, and the long future prediction period from $t=60~\text{to}~364$ days.
  \textit{(a):} \textit{Top} plots show the yearly variation of solar radiation and the \textit{bottom} plots the aggregated states from the NNPZD model (\textit{ground truth}) overlaid with the dynamic mixed layer depth in \textit{dashed red} lines.  
   In the subsequent plots \textit{(b)}, \textit{(c)}, \textit{(d)}, and \textit{(e)}, we show the absolute difference of the different neural closure cases with the ground truth.
   For each case, we also provide the pair of time-averaged errors (see Fig.\;\ref{fig: Exp2 Comparison of high and low-res solutions} for description).
   These results correspond to the architectures given in Table \ref{table: Exp3 and 4 architecture}.
    %
  }
  \label{fig: Exp4 Main results comparison between nODE, nDDE and nDistDDE}
\end{figure}

For our final set of experiments, we extend the ODE models used in Experiments-3a (Sec.~\ref{sec:biogeochemical 0-D models}) to contain a vertical dimension (thus, 1-D) and vertical eddy mixing parameterized by the operator, $\partial/\partial z \left (K_z(z, M)\partial/\partial z (\bullet) \right)$, where $K_z$ is a dynamic eddy diffusion coefficient. 
A mixed layer of varying depth ($M = M(t)$) is used as a physical input to the ecosystem models. Thus, each biological state variable $B(z, t)$ is governed by the following non-autonomous PDE,
\begin{equation}
\label{eq: ADR eqn}
    \frac{\partial B}{\partial t} = S^B + \frac{\partial}{\partial z}\left( K_z(z, M(t))\frac{\partial B}{\partial z}\right) \;,
\end{equation}
where $S^B$ are the corresponding biology source terms, 
which also makes it stiff.
The dynamic depth-dependent diffusion parameter $K_z$ is given by,
\begin{equation}
    K_z(z, M(t)) = K_{z_b} + \frac{(K_{z_0} - K_{z_b})(\arctan(-\gamma (M(t) - z)) - \arctan(-\gamma (M(t) - D)))}{\arctan(-\gamma M(t)) - \arctan(-\gamma (M(t) - D))} \; ,
\end{equation}
where $K_{z_b}$ and $K_{z_0}$ are the diffusion at the bottom and surface respectively, $\gamma$ is the thermocline sharpness, and $D$ is the total depth.
The 1-D model and parameterizations are adapted from Eknes and Evensen, 2002 \cite{eknes2002ensemble}, and Newberger et. al., 2003 \cite{newberger2003analysis}. They simulate the seasonal variability in upwelling, sunlight, and  biomass vertical profiles. 
The dynamic mixed layer depth, 
surface photosynthetically-available radiation $I_0(t)$, and biomass fields $B(z, t)$ are shown in Figure~\ref{fig: Exp4_npz_agg_sol_modified}. The radiation $I_0(t)$ and total biomass concentration, $T_{bio}(z,t)$, affect $S^B$ and the initial conditions.

For these Experiments-3b, we consider $20$ grid points in the vertical and use the \textit{dopri5} \cite{hairer1993solving} scheme with adaptive time-stepping.
Data is stored at every time-step of $\Delta t = 0.1~days$ for all our time-integration requirements. Solutions of aggregated states of the high-complexity 1-D NNPZD model (true data) and their absolute difference with the corresponding low-complexity 1-D NPZ model states are provided in Figs.\;\ref{fig: Exp4_npz_agg_sol_modified} \& \ref{fig: Exp4_npz_nnpzd_diff_modified}, respectively. 
For training our neural closure models for the 1-D NPZ model, we use the same training regiment as in Experiments-1, 2, \& 3a (Secs.\;\ref{sec: exp1}, \ref{sec: exp2}, \& \ref{sec:biogeochemical 0-D models}). 
We note that in the 1-D NPZ model, the local mixing across depths occurs only due to the eddy diffusion term, and not to the biology source terms. Thus, we employ 1-D convolutional layers with receptive fields of size 1. We again use the custom layer, \textit{BioConstrainLayer} (Sec.\;\ref{sec:biogeochemical 0-D models}), to ensure that the sum of the biology source terms of the augmented 1-D NPZ model does not leave any new residual due to the neural closure terms. Along with this, we define a new custom layer, called \textit{AddExtraChannels}, to add additional channels to the input of this layer. We add one for the depths at different grid points, and the other for the corresponding values of available sunlight for photo-chemical reactions ($I(z, t)$). The architectures details for the three closure models used are presented in Table\;\ref{table: Exp3 and 4 architecture}. For the nODE, we again employ a bigger architecture, and for the  discrete-nDDE, only four discrete delay values are used. 
The values of other hyperparameters are given in Sec.\;\ref{SI: hyperparameters}. The training period ranges from $t= 0$ to $30~days$ and validation period from $t= 30$ to $60~days$, both within the first season. The prediction period, however, ranges from $t= 60$ to $364~days$: it is more than 10 times longer than the training period and involves the four seasons. Together, the three periods span a full year. For loss function, we combine the $L_2$ errors, considering all the biological states computed for individual depths, and then averaged over all the depths and times, $ \mathcal{L} = \allowdisplaybreaks  \frac{1}{M} \sum_{i=1}^M \left(\frac{1}{N_z}\sum_{k=1}^{N_z = 20}\left(\sqrt{\sum_{B\in\{N, P, Z\}} |B^{pred}(z_k, T_i) - B^{true}(z_k, T_i)|^2}\right) \right) $.
%

The performance of the three neural closure models augmenting the 1-D NPZ model is evaluated after 200 epochs of training (the stochastic gradient descent nearly converges, see Fig.\;\ref{fig: Exp_4_train_val_loss_all}). The truth fields are the aggregated biology variables from the high-complexity 1-D NNPZD model (Eqs.\;\ref{eq:NNPZD Model} \& \ref{eq: ADR eqn}) spanning training, validation, and prediction periods. Results are presented in Figure \ref{fig: Exp4 Main results comparison between nODE, nDDE and nDistDDE}. 
We find again that despite the bigger architecture for the nODE case, it develops spurious oscillations around $t = 250~days$. The discrete-nDDE and distributed-nDDE, both with smaller architectures, however match well with the true variables for nearly the full year of simulation, about 10 months of which is future prediction. The distributed-nDDE performs slightly better than its counterpart.
In Fig.\;\ref{fig: Exp4 Main results comparison between nODE, nDDE and nDistDDE}, we also provide averaged error numbers for the baseline (Fig.\;\ref{fig: Exp4_npz_nnpzd_diff_modified}) and the different closure models, all of which improve the baseline.
As in Experiments-3a, we again notice large spikes in the starting of the training regime, for the same reason as given earlier, and similar trends for hyperparameter sensitivity. We also found that the Experiments-3b were affected by the choice of loss function. For example, using $L_2$ error computed for each biological state vector (containing values for all the depths) and then averaging over the number of biological states and times deteriorated the quality of learning.   

Despite the presence of complex physical processes and relatively large dimensions as compared to the previous experiments, the nDDEs closures were found to effectively match the high-complexity model and maintain long-term accuracy.

\subsection{Computational Complexity}
It is crucial to analyze complexity and in particular the cost of adding a neural closure model to a low-fidelity model. In this section, we analyze the computational complexity in terms of \textit{flop} (floating point operations) count for evaluating the right-hand-side (RHS) of the low-fidelity models, and for the forward-pass of the neural closure models \cite{mizutani2001complexity}. We will also comment on the training costs.  
%
%
The Burger's PDE considered for Experiments-1\&2 (Secs.\;\ref{sec: exp1}~\&~\ref{sec: exp2}) has a nonlinear advection term. Hence, for the POD-GP ROM and the FOM, the upper \textit{flops} is of the order of the square of number of resolved modes and of the spatial grid resolution, respectively. 
%
In general, for reaction terms and biogeochemical systems, the number of nonlinear parameterizations present are of the order of the number of components in the model. Hence, even for Experiments-3a (Sec.\;\ref{sec:biogeochemical 0-D models}), the upper \textit{flops} is of the order of the square of the number of biological components. For Experiments-3b (Sec.\;\ref{sec:biogeochemical 1-D models}), the upper \textit{flops} is also affected by the diffusion terms.
%
Let the number of state variables in the low-fidelity models be $N \in \mathbb{N}$, thus the leading order computational complexity would be $\mathcal{O}(cN^2)$, where $c \in \mathbb{R}^+$ is some constant dependent on the numerical schemes used for spatial discretization, the exact functional form of the RHS, etc.

Now, when neural closure models are added to the low-fidelity models, the time integration requires a forward pass through the neural-network. 
%
This cost varies with the neural architecture and model type, here either a fully-connected or convolutional, and discrete-nDDE or distributed-nDDE, respectively. 
%
As observed in our experiments, using delays in the closure model enables us to use shallower networks, with a depth independent of the number of state variables ($N$). We also found that the width of the networks was similar to, or smaller than, $N$.
In case of distributed-nDDEs, we observed that
the width of the auxiliary network ($g_{NN}$) could be on an average nearly half the size of the main network ($f_{NN}$).
Let the size of the hidden state for the RNN in discrete-nDDEs be $N_h \in \mathbb{N}$, and the number of neurons in the hidden layers of the main and auxiliary networks in the case of distributed-nDDEs be $N_h$ and $N_h/2$, respectively, with $N_h \lesssim N$. It could be easily shown that the leading order complexity for a single iteration of RNN would be $\mathcal{O}(N_h^2 + N_hN)$, which is due to the hidden and input state vectors being multiplied by the weight matrices, while the application of activation function would be $\mathcal{O}(N_h)$ only. As the number of discrete-delays in discrete-nDDEs are independent of $N$ and $\mathcal{O}(1)$, it does not affect the complexity of the RNN. The complexity of the first hidden layer and/or the output layer of the deep neural-networks used in discrete-nDDEs and distributed-nDDEs (main network, $f_{NN}$) will be $\mathcal{O}(N_hN)$ and $\mathcal{O}(N_h(N + N_h/2))$, respectively, while for the remaining hidden layers, it will be $\mathcal{O}(N_h^2)$.
Focusing on the integral of the auxiliary network ($g_{NN}$) over the delay period in distributed-nDDEs, if implemented efficiently, at every time-step,
we only need to compute 
the integral twice over periods of size $\Delta t$, each adjacent to the ends of the present delay-period. We can add and subtract these integrals over $\Delta t$ periods to compute the overall integral in a rolling window sense. Hence, the contribution to the computational cost by the auxiliary network would be $\mathcal{O}(N_h N/2)$ (for first hidden layer) and $\mathcal{O}(N_h^2/4)$ (subsequent hidden layers). Considering $D \in \mathbb{N}$ as the depth for all of the networks considered, the complexity for the forward pass through the discrete-nDDE closure is $\mathcal{O}(DN_h^2 + DN_hN)$, while for distributed-nDDE, it is $\mathcal{O}((3/2)N_hN + (7/4)DN_h^2)$. 
%
These costs were computed considering fully-connected layers, however, will only be cheaper in case of convolutional layers.
%
Thus, the additional computational cost due to the presence of neural closure models is of similar or lower complexity than the existing low-fidelity model. 

Estimating the computational cost/complexity of training in \textit{flops} is not common because apart from time-integrating the forward model and adjoint equations, there are many other operations such as here: automatic differentiation through the neural networks; creation and use of interpolation functions; the integral to compute the final derivatives; the gradient descent step, etc. The overall cost also depends on the number of epochs needed for convergence. The present training cost is of course non-negligible, as with any supervised learning algorithm. However, in applications where one needs to repeatedly solve a low-fidelity model, investing in a one-time cost of training a neural closure model can later lead to accuracy close to that of the high-fidelity model with only a small increase in the computational cost of the low-fidelity model.


\section{Conclusion}
We developed a novel, versatile, rigorous, and unified methodology to learn closure parameterizations for low-fidelity models using data from high-fidelity simulations. The Mori-Zwanzig formulation \cite{stinis2015renormalized, chorin2000optimal, gouasmi2017priori} and the presence of inherent delays in 
complex dynamical systems \cite{erneux2009applied}, especially biological systems \cite{glass2020nonlinear, tokuda2019reducing},
justify the need for non-Markovian closure parameterizations. To learn such non-Markovian closures, our new \textit{neural closure models} extend neural ordinary differential equations (nODEs; \cite{chen2018neural}) to neural delay differential equations (nDDEs).
Our nDDEs do not require access to the high-fidelity model or frequent enough and uniformly-spaced high-fidelity data to compute the time derivative of the state with high accuracy.
Further, it enables the accounting of errors in the time-evolution of the states in the presence of neural networks during training. 
We derive the adjoint equations and network architectures needed to efficiently implement the nDDEs, for both discrete and distributed delays, agnostic to the specifics of the time-integration scheme, and capable of handling stiff systems.
For distributed-delays, we propose a novel architecture consisting of two coupled deep neural networks, which enables us to incorporate memory without the use of any recurrent architectures.

Through simulation experiments, we showed that our methodology drastically improves the 
long-term
predictive capability of low-fidelity models 
for the main classes of model truncations.
Specifically, our neural closure models efficiently account for truncated modes in reduced-order-models (ROMs), capture the effects of subgrid-scale processes in coarse models, and augment the simplification of complex 
biological
and non-autonomous physical-biogeochemical models.
Our first two classes of simulation experiments utilize the advecting shock problem governed by the Burger's PDE, with its low-fidelity models derived either by proper-orthogonal-decomposition Galerkin projection or by reducing the spatial grid resolution. 
Our third class of experiments considers marine biological ODEs of varying complexities and their physical-biogeochemical PDE extensions with non-autonomous dynamic parameterizations. 
The low-fidelity models are obtained by aggregation of components and other simplifications of processes and parameterizations.
In each of these classes, results consistently show that using non-Markovian over Markovian closures improves the accuracy of the learned system while also requiring smaller network architectures.
Our use of the known-physics/low-fidelity model also helps to reduce the required size of the network architecture and the number of time samples for the training data.
We also outperform classic dynamic closures such as the Smagorinsky subgrid-scale model. 
These results are obtained using stringent evaluations: we compare the performance of the learned system for the training period (during which high-fidelity data snapshots are used for training) and validation period (during which hyperparameter tuning occurs) as often done, but we also compare it for much longer-term future prediction periods with no overlap with the preceding two. We even consider a prediction period reaching $10$ times the length of the training/validation period, thus successfully demonstrating the extrapolation capabilities of nDDE closures.

In our experiments, we find that just using a few numbers of discrete delays might perform equally well or better than using a distributed delay which involves an integral of the state variable over a delay period. We provide a plausible explanation of this counter-intuitive observation using the data processing inequality from information theory. 
We also show that there exists an optimal amount of past information
to incorporate for a specified architecture and the relevant time-scales present in the dynamical system,
thus indicating that neither too little nor too much past information is helpful. Finally, a computational complexity analysis using \textit{flop} (floating point operation) count proves that the additional computational cost due to the presence of our neural closure models is of similar or lower complexity than the existing low-fidelity model.

The present work provides a unified framework to learn non-Markovian closure parameterization using delay differential equations and neural networks.
It enables the use of the often elusive Mori-Zwanzig formulation \cite{stinis2015renormalized, chorin2000optimal, gouasmi2017priori} in its full glory without unjustified approximations and simplifications. Our nDDE closures are not just limited to the shown experiments, but could be widely extended to other fields such as control theory, robotics, pharmacokinetic-pharmacodynamics, chemistry, economics, biological regulatory systems, etc.

\vspace{0.75cm}


\noindent\textbf{Data Accessibility. }{The codes and data used in this work are available in the GitHub repository: \url{https://github.com/mit-mseas/neuralClosureModels.git}}
%

\vspace{0.25cm}

\noindent\textbf{Author's Contributions. }{A.G. conceived the idea of using neural delay differential equations for closure parameterizations; derived the adjoint equations; implemented the neural network architectures and the simulation experiments; interpreted the computational results; and wrote a first draft of the manuscript. P.F.J.L. supervised the work; conceived the ideas to extend the methodology to capture the effects of subgrid-scale processes in coarse models, and to augment the simplification of  complex mathematical models; interpreted the computational results; and edited and wrote significant parts of the manuscript.}

\vspace{0.25cm}

\noindent\textbf{Competing Interests. }{We declare we have no competing interests.}

\vspace{0.25cm}

\noindent\textbf{Funding. }{We are grateful to the Office of Naval Research for partial support under
grant N00014-20-1-2023 (MURI ML-SCOPE) to the
Massachusetts Institute of Technology. We also thank MathWorks and the Mechanical Engineering Department at MIT for awarding a competitive 2020-2021 MathWorks Mechanical Engineering Fellowship for A.G. Some of the computations were made possible due the Google Cloud Platform research credits, which we gratefully acknowledge.}

\vspace{0.25cm}

\noindent\textbf{Acknowledgement. }{We thank the members of our MSEAS group for their collaboration and insights, especially Mr.~Aaron Charous. 
We thank our ML-SCOPE team for many useful discussions, especially Dr. Micka\"el Chekroun for his comments on the final draft of the manuscript.
We also thank the anonymous reviewers for their constructive feedback which helped improve the manuscript.
%
}




\bibliographystyle{abbrvnat}
\bibliography{mseas,references}

\makeatletter\@input{msxx.tex}\makeatother
\end{document}


\date{\today}
\maketitle

\section{Mori-Zwanzig Formulation}
\label{sec: MZ formulation in supp info}
Without loss of generality, the full nonlinear dynamical system model is written as,
\begin{equation}
        \frac{d u_k(t)}{dt} = R_k(u(t), t) \, , \quad \text{with}\quad u_k(0) = u_{0k}\,, \quad k \in \mathfrak{F} \,.
    \label{eq: very general dynamical system}
\end{equation}
The full state vector is  $u = (\{ u_k \}), \; k \in \mathfrak{F} = \mathfrak{R} \cup \mathfrak{U}$, where $\mathfrak{R}$ is the set corresponding to the resolved variables (e.g.\;coarse field or reduced variables), and $\mathfrak{U}$ the set corresponding to the unresolved variables  (e.g.\;subgrid field or complement variables), which as a union, $\mathfrak{F}$, form the set for full space of variables. We also denote $u = \{\hat{u}, \tilde{u}\}$ where $\hat{u} = (\{ u_k \}), \; k \in \mathfrak{R}$ and $\tilde{u} = (\{ u_k \}),\; k \in \mathfrak{U}$. Similarly, $u_0 = \{\hat{u}_0, \tilde{u}_0\} $, with $\hat{u}_0 = (\{ u_{0k} \}), \; \in \mathfrak{R}$ and $\tilde{u}_0 = (\{ u_{0k} \}), \;  k \in \mathfrak{U}$.

We can write the above non-linear system of ODEs (Eq.\;\ref{eq: very general dynamical system}) exactly as a system of linear PDEs by casting it in the Liouville form,
\begin{equation}
        \qquad\frac{\partial \phi_k}{\partial t} = L\phi_k \,, \quad \text{with} \quad
        \phi_k(u_0, 0) = u_{0k}
\,, \quad k \in \mathfrak{F}\,,
    \label{eq:linear PDE system}
\end{equation}
where the Liouville operator is $L = \sum_{ i \in \mathfrak{F}} R_i(u_0)\frac{\partial}{\partial u_{0i}}$, with $R_i$ denoting element $i$ of the full model dynamics (Eq.\;\ref{eq: very general dynamical system}). The solution of Eq.\;\ref{eq:linear PDE system} is given by $u_k(u_0, t) = \phi_k(u_0, t) = e^{tL} \phi_k(u_0, 0)$. Hence, we can also rewrite Eq.\;\ref{eq:linear PDE system} as,
\begin{equation}
    \begin{split}
        \frac{\partial }{\partial t}e^{tL} u_{0k} = Le^{tL} u_{0k} \,, \qquad k \in \mathfrak{F} \,.
    \end{split}
    \label{eq: rewrite linear PDE system}
\end{equation}
Now, let $P$ be a orthogonal projection on the space of functions of the resolved initial conditions $\hat{u}_0$, such that, for any nonlinear function $h(u_0) = h(\{\hat{u}_0, \tilde{u}_0\})$, then $P(h(u_0)) = h(\hat{u}_0)$. Similarly, $Q = I - P$ is the projection on the null space of $P$. It is important to note that the projectors $P$  and $Q$ used in this formulation are fundamentally different from $L_2$~projectors. Using the Dyson's formula $e^{tL} = e^{tQL} + \int_0^t e^{(t-s)L} PL e^{sQL} ds$, and noting that $L$ and $e^{tL}$ commute, we can then exactly rewrite Eq.\;\ref{eq: rewrite linear PDE system} as, 
\begin{equation}
    \begin{split}
        \frac{\partial }{\partial t}e^{tL} u_{0k} = e^{tL} PL u_{0k} + e^{tQL}QLu_{0k} + \int_0^t e^{(t-s)L} PL e^{sQL} QL u_{0k} ds \, , \quad k \in \mathfrak{R}\,,
    \end{split}
    \label{eq: rewrite linear PDE system in MZF}
\end{equation}
which is called the Mori-Zwanzig (MZ) formulation. Importantly, the above equation is an exact representation of Eq.\;\ref{eq: rewrite linear PDE system} for the resolved components. For convenience, we denote $F_k(u_0, t) = e^{tQL} QL u_{0k}$ and $K_{k}(u_0, t) = PLF_k(u_0, t)$, and thus further rewrite Eq.\;\ref{eq: rewrite linear PDE system in MZF} as,
\begin{equation}
    \begin{split}
        \frac{\partial }{\partial t}u_k(u_0, t) = \underbrace{R_k(\hat{u}(u_0, t))}_{\text{Markovian}} + \underbrace{F_k(u_0, t)}_{\text{Noise}} + \underbrace{\int_0^t K_k(\hat{u}(u_0, t - s), s) ds}_{\text{Memory}} , \quad k \in \mathfrak{R}\,,
    \end{split}
    \label{eq: rewrite linear PDE system in MZF concise}
\end{equation}
where 
$R_k$ is again the same as that in the full model dynamics given by Eq.\;\ref{eq: very general dynamical system}.
Eq.\;\ref{eq: rewrite linear PDE system in MZF concise} provides useful guidance for closure modeling. The first term in Eq.\;\ref{eq: rewrite linear PDE system in MZF concise} is the Markovian term dependent only on the values of the variables at the present time, while the closure consists of two terms: the noise term and a memory term that is non-Markovian. 
We can further simplify Eq.\;\ref{eq: rewrite linear PDE system in MZF concise} by applying the $P$ projection and using the fact that the noise term lives in the null space of $P$ for all times, which could be easily proved. For ROMs with initial conditions devoid of any unresolved dynamics, i.e.\;$\tilde{u}_0 = 0$ and thus $u_0 = \hat{u}_0$, we then retain the exact dynamics after the projection step, noticing in this case that $Pu_k(u_0, t) = u_k(\hat{u}_0, t), \forall k \in \mathfrak{R}$,
\begin{equation}
    \begin{split}
        \frac{\partial }{\partial t}u_k(\hat{u}_0, t) = P R_k(\hat{u}(\hat{u}_0, t)) + P\int_0^t K_k(\hat{u}(\hat{u}_0, t - s), s) ds , \quad k \in \mathfrak{R}\,.
    \end{split}
    \label{eq: MZF with no unresolved ICs}
\end{equation}
Hence, for such systems, the closure model would only consider the non-Markovian memory term. The above derivation of the MZ formulation has been adapted from \cite{stinis2015renormalized,wang2020recurrent,gouasmi2017priori}.

\section{Adjoint Equations for Neural Delay Differential Equations}
\label{app: adjoint equations}
Here, we provide a detailed derivation of adjoint equations for neural DDEs with discrete and distributed delays. For related derivations, we refer to  \cite{bradley2013pde, calver2017numerical}.

\subsection{Discrete-nDDE}
The neural-network parameterized discrete DDE is given by,
\begin{equation}
\begin{split}
\frac{du(t)}{dt} &= f_{RNN}(u(t), u(t - \tau_1), ..., u(t - \tau_K), t; \theta), \quad t \in (0, T] \\
u(t) &= h(t), \quad t \leq 0
\end{split}
\end{equation}
where $\tau_1, ..., \tau_K$ are $K$ number of discrete delays and $f_{RNN}(\bullet, t; \theta)$ is any recurrent architecture with trainable parameters $\theta$. Let data be available at $M$ times, $T_1 < ... <T_M \leq T$. 
%
Our goal is to optimize the total loss function, $\mathcal{L} = \int_0^T \sum_{i=1}^M l(u(t))\delta(t - T_i)dt$ (where $l(\bullet)$ are scalar loss functions such as mean-squared-error (MSE), and $\delta(t)$ is the Kronecker delta function), given the data and nDDE.
%

We first start by writing the Lagrangian for the above system,
\begin{equation}
\begin{split}
   L =& \mathcal{L}(u(t)) + \int_0^T \lambda^T(t) \left( d_t{u}(t) - f_{RNN}(u(t), u(t-\tau_1), ..., u(t-\tau_K), t; \theta)\right)dt \\
   & + \int_{-\tau_K}^0 \mu^T(t)(u(t) - h(t))dt
   \end{split}
\end{equation}
where $\lambda(t)$ and $\mu(t)$ are the Lagrangian variables, and where, for brevity, we use $\frac{\partial}{\partial (\bullet)} \equiv \partial_{(\bullet)}$ and $\frac{d}{d (\bullet)} = d_{(\bullet)}$ from now on. We also assume that the loss function ($\mathcal{L}$) and the initial conditions ($h(t),~t\leq 0$) are independent of $\theta$. Hence, the derivative of the Lagrangian w.r.t. $\theta$ is given by, 
\begin{equation}
\label{eq: derivative of lagrangian}
\begin{split}
   d_{\theta} L =& \partial_{u(t)} \mathcal{L}(u(t)) d_{\theta}u(t) + \int_0^T \lambda^T(\dt d_t u(t) - \partial_{u(t)}f_{RNN}(\bullet, t; \theta)\dt u(t)   \\
   & -  \partial_{u(t-\tau_1)}f_{RNN}(\bullet,t; \theta)\dt u(t-\tau_1) ... - \partial_{u(t-\tau_K)}f_{RNN}(\bullet,t; \theta)\dt u(t-\tau_K) - \partial_{\theta}f_{RNN}(\bullet,t; \theta)) dt .
   \end{split}
\end{equation}
Using integration-by-parts, we can write,
\begin{equation}
   \int_0^T \lm^T \dt d_t u(t) dt = \lm^T(T)\dt u(T) - \lm^T(0)\dt u(0) - \int_0^T d_t{\lm}^T(t) \dt u(t) dt
\end{equation}
and by change of variables,
\begin{equation}
\begin{split}
    \int_0^T & \lm^T(t)  \partial_{u(t - \tau_i)}f_{RNN}(u(t), u(t-\tau_1), ..., u(t-\tau_K), t; \theta) \dt u(t - \tau_i) dt  \\
   =& \int_{-\tau_i}^{T - \tau_i} \lm^T(t+\tau_i)\partial_{u(t)}f_{RNN}(u(t+\tau_i), u(t-\tau_1+\tau_i), ..., u(t-\tau_K+\tau_i), t + \tau_i; \theta) \dt u(t) dt  \\
   =& \int_{0}^{T} \lm^T(t+\tau_i)\partial_{u(t)}f_{RNN}(u(t+\tau_i), u(t-\tau_1+\tau_i), ..., u(t-\tau_K+\tau_i), t + \tau_i; \theta) \dt u(t) dt \\
   &+ \int_{-\tau_i}^{0} \lm^T(t+\tau_i)\partial_{u(t)}f_{RNN}(u(t+\tau_i), u(t-\tau_1+\tau_i), ..., u(t-\tau_K+\tau_i), t + \tau_i; \theta) \dt u(t) dt  \\
   & - \int_{T-\tau_i}^{T} \lm^T(t+\tau_i)\partial_{u(t)}f_{RNN}(u(t+\tau_i), u(t-\tau_1+\tau_i), ..., u(t-\tau_K+\tau_i), t + \tau_i; \theta) \dt u(t) dt \\
   =& \int_{0}^{T} \lm^T(t+\tau_i)\partial_{u(t)}f_{RNN}(u(t+\tau_i), u(t-\tau_1+\tau_i), ..., u(t-\tau_K+\tau_i), t + \tau_i; \theta) \dt u(t) dt  \\
   & + \int_{-\tau_i}^{0} \lm^T(t+\tau_i)\partial_{u(t)}f_{RNN}(u(t+\tau_i), u(t-\tau_1+\tau_i), ..., u(t-\tau_K+\tau_i), t + \tau_i; \theta) \dt u(t) dt \,.
   \end{split}
   \label{eq: chnage of variables nDDE-RNN}
\end{equation}
We further assume $\lm (t) = 0, ~t\geq T$. Inserting everything back into Eq.\;\ref{eq: derivative of lagrangian}, we obtain,
\begin{equation}
\begin{split}
   d_{\theta} L =& \int_0^T\left(\sum_{i = 1}^M\partial_{u(t)} l(u(t))\delta(t - T_i) - d_t \lm^T(t) - \lm^T(t)\partial_{u(t)}f_{RNN}(u(t), u(t - \tau_1), ..., u(t - \tau_K), t; \theta) \right. \\
   & \left.- \sum_{i = 1}^K \lm^T(t+\tau_i)\partial_{u(t)}f_{RNN}(u(t+\tau_i), u(t-\tau_1+\tau_i), ..., u(t-\tau_K+\tau_i), t + \tau_i; \theta)\right) \dt u(t) dt \\
   &- \int_0^T \lambda^T(t)\partial_{\theta}f_{RNN}(u(t), u(t - \tau_1), ..., u(t - \tau_K), t; \theta)dt + \lm^T(T)\dt u(T) - \lm^T(0)\dt u(0) \\
   & + \sum_{i = 1}^K \int_{-\tau_i}^{0} \lm^T(t+\tau_i)\partial_{u(t)}f_{RNN}(u(t+\tau_i), u(t-\tau_1+\tau_i), ..., u(t-\tau_K+\tau_i), t + \tau_i) \dt u(t) dt \;.
   \end{split}
\end{equation}
The last two term in the above equation is zero because of the user-defined initial condition are independent of $\theta$. Further, we aim to eliminate of $\dt u(t)$ everywhere, because avoiding the need to compute it explicitly is the main premise of the adjoint method. We can do this if we assume,
\begin{equation}
\begin{split}
   d_t\lm^T(t) =& \sum_{i = 1}^M\partial_{u(t)} l(u(t))\delta(t - T_i) - \lm^T(t)\partial_{u(t)}f_{RNN}(u(t), u(t - \tau_1), ..., u(t - \tau_K), t; \theta) \\
   & - \sum_{i = 1}^K \lm^T(t+\tau_i)\partial_{u(t)}f_{RNN}(u(t+\tau_i), u(t-\tau_1+\tau_i), ..., u(t-\tau_K+\tau_i), t + \tau_i; \theta), \; t \in [0, T) \\
   \lm(t) =& 0, \qquad t\geq T
   \end{split}
\end{equation}
Finally, after solving the above adjoint equation in $\lm(t)$, we can compute the required derivative $\dt L$ as,
\begin{equation}
   \dt L = - \int_0^T \lambda^T(t)\partial_{\theta}f_{RNN}(u(t), u(t - \tau_1), ..., u(t - \tau_K), t; \theta)dt \;.
\end{equation}

\subsection{Distributed-nDDE}
The neural network parameterized distributed DDE is given by,
\begin{equation}
\begin{split}
\frac{du(t)}{dt} &= f_{NN}\left(u(t), \int_{t-\tau_2}^{t-\tau_1} g_{NN}(u(\tau), \tau; \phi)d\tau, t; \theta \right), \quad t \in (0, T] \\
u(t) &= h(t), \quad t \leq 0
\end{split}
\end{equation}
where $\tau_2 \geq \tau_1$ are the delay amounts; and $f_{NN}(\bullet; \theta)$, $g_{NN}(\bullet; \phi)$ are neural-networks with trainable parameters $\theta$, $\phi$ respectively. By introducing a new variable $y(t) = \int_{t-\tau_2}^{t-\tau_1} g_{NN}(u(\tau), \tau; \phi)d\tau$, we can rewrite the above equation as a coupled discrete DDEs,
\begin{equation}
\begin{split}
\frac{du(t)}{dt} &= f_{NN}\left(u(t), y(t), t; \theta \right), \quad t \in (0, T] \\
\frac{dy(t)}{dt} &= g_{NN}(u(t-\tau_1), t-\tau_1; \phi) - g_{NN}(u(t-\tau_2), t-\tau_2; \phi), \quad t \in (0, T] \\
u(t) &= h(t), \quad \tau_2 \leq t \leq 0\\
y(0) & = \int_{-\tau_2}^{-\tau_1} g_{NN}(h(t), t; \phi)dt
\end{split}
\end{equation}
Also, let data be available at $M$ times, $T_1 < ... <T_M \leq T$. Our goal is to optimize the scalar loss function, $\mathcal{L}(u(t)) = \int_0^T \sum_{i=1}^M l(u(t))\delta(t - T_i)dt$, given the data and nDDE. 

We will again start by writing the Lagrangian for this setup,
\begin{equation}
\begin{split}
   L =& \mathcal{L}(u(t)) + \int_0^T \lm^T(t) (d_t u(t) - f_{NN}(u(t), y(t), t; \theta))\, dt \\
   & + \int_0^T \mu^T(t) \left({ d_t y(t) - g_{NN}(u(t - \tau_1), t-\tau_1; \phi)  + g_{NN}(u(t - \tau_2), t-\tau_2; \phi) }\right)dt  \\
   & + \int_{-\tau_2}^0 \gamma^T(t)(u(t) - h(t) ) dt  + \alpha^T \left (y(0) - \int_{-\tau_2}^{-\tau_1} g_{NN}(h(t), t; \phi)dt \right) \,, 
   \end{split}
\end{equation}
where $\lambda(t), \mu(t), \gamma(t)$, and $\alpha$ are the Lagrangian variables.
Now in this case, we need to obtain the derivatives of the Lagrangian w.r.t. both $\theta$ and $\phi$. We will first obtain $\dt L$,
\begin{equation}
\begin{split}
   \dt L =& \partial_{u(t)} \mathcal{L}(u(t))\dt u(t) + \int_0^T \lm^T(t) \left(\dt d_t u(t) - \partial_{u(t)} f_{NN}(u(t), y(t), t; \theta)\dt u(t) \right. \\
   & \left. - \partial_{y(t)} f_{NN}(u(t), y(t), t;\theta) \dt y(t) - \partial_{\theta}f_{NN}(u(t), y(t), t; \theta)\right)dt   \\
   & + \int_0^T \mu^T(t) \left(\dt d_t y(t)  - \partial_{u(t - \tau_1)}g_{NN}(u(t - \tau_1), t-\tau_1; \phi)\dt u(t - \tau_1) \right. \\
   & \left. + \partial_{u(t-\tau_2)}g_{NN}(u(t-\tau_2), t-\tau_2; \phi)\dt u(t - \tau_2)\right)dt  \,.
   \end{split}
   \label{eq: derivative lagrangian nDDE-Dist}
\end{equation}
Using integration-by-parts, we can write,
\begin{equation}
\begin{split}
   \int_0^T \lm^T(t) \dt d_t u(t) dt =& \lm^T(T) \dt u(T) - \lm^T(0)\dt u(0) - \int_0^T d_t\lm^T(t) \dt u(t) dt \\
   \int_0^T \mu^T(t) \dt d_t y(t) dt =& \mu^T(T) \dt y(T) - \mu^T(0)\dt y(0) - \int_0^T d_t\mu^T(t) \dt y(t) dt
   \end{split}
   \label{eq: integration by parts nDDE-Dist w.r.t theta}
\end{equation}
and by change of variables,
\begin{equation}
\begin{split}
    \int_0^T \mu^T(t) \partial_{u(t - \tau_i)}g_{NN}(u(t - \tau_i), t-\tau_i; \phi) & \dt u(t - \tau_i) dt  \\
   =& \int_{-\tau_i}^{T - \tau_i} \mu^T(t+\tau_i)\partial_{u(t)}g_{NN}(u(t), t; \phi) \dt u(t) dt  \\
   =& \int_{0}^{T} \mu^T(t+\tau_i)\partial_{u(t)}g_{NN}(u(t), t; \phi) \dt u(t) dt \\
   &+ \int_{-\tau_i}^{0} \mu^T(t+\tau_i)\partial_{u(t)}g_{NN}(u(t), t; \phi) \dt u(t) dt  \\
   & - \int_{T-\tau_i}^{T} \mu^T(t+\tau_i)\partial_{u(t)}g_{NN}(u(t), t; \phi) \dt u(t) dt \\
   =& \int_{0}^{T} \mu^T(t+\tau_i)\partial_{u(t)}g_{NN}(u(t), t; \phi) \dt u(t) dt  \\
   & + \int_{-\tau_i}^{0} \mu^T(t+\tau_i)\partial_{u(t)}g_{NN}(u(t), t; \phi) \dt u(t) dt
   \end{split}
   \label{eq: chnage of variables nDDE-Dist}
\end{equation}
We further assume $\mu(t) = 0, ~t\geq T$. Inserting everything back (Eqs.\;\ref{eq: integration by parts nDDE-Dist w.r.t theta} and \ref{eq: chnage of variables nDDE-Dist}) into Eq.\;\ref{eq: derivative lagrangian nDDE-Dist}, and keeping in mind that the initial condition $h(t)$ is independent of $\theta$, we obtain,
\begin{equation}
\begin{split}
   d_{\theta} L =& \int_0^T\left(\sum_{i = 1}^M\partial_{u(t)} l(u(t))\delta(t - T_i) - d_t\lm^T(t) - \lm^T(t)\partial_{u(t)}f_{NN}(u(t), y(t), t; \theta) \right. \\
  &  - \mu^T(t + \tau_1) \partial_{u(t)}g_{NN}(u(t), t; \phi) +  \mu^T(t+\tau_2) \partial_{u(t)}g_{NN}(u(t), t; \phi)\bigg) \dt u(t) dt  \\
  & - \int_0^T \lm^T(t) \partial_{\theta}f_{NN}(u(t), y(t), t; \theta) dt \\
   & + \int_0^T (-d_t\mu^T(t) - \lm^T(t)\partial_{y(t)}f_{NN}(u(t), y(t), t; \theta))\dt y(t) dt  \\
   & + \lm^T(T)\dt u(T) + \mu^T(T)\dt y(T) \;.
   \end{split}
\end{equation}
Again, the objective is to avoid the need to compute $\dt u(t)$ and $\dt y(t)$, hence we assume, $\mu(t) = 0, ~t\geq T$; and $\lm(T) = 0$. We can write the following coupled adjoint equations,
\begin{equation}
\label{eq: adjoint equations nDDE-Dist}
\begin{split}
   d_t\lm^T(t) =& \sum_{i = 1}^M\partial_{u(t)} l(u(t))\delta(t - T_i) - \lm^T(t)\partial_{u(t)}f_{NN}(u(t), y(t), t; \theta) \\
   & - \mu^T(t+\tau_1) \partial_{u(t)}g_{NN}(u(t), t; \phi) \\
   & +  \mu^T(t+\tau_2) \partial_{u(t)}g_{NN}(u(t), t; \phi)\,, \qquad t \in [0, T) \\
   d_t\mu^T(t) =& - \lm^T(t)\partial_{y(t)}f_{NN}(u(t), y(t), t; \theta)\,, \qquad t \in [0, T) \\
   \lm^T(t)  =& ~ 0 \quad \text{and} \quad \mu^T(t) = 0, \qquad t \geq T \;.
   \end{split}
\end{equation}
Finally after solving the coupled adjoint equations in $\lm(t)$ and $\mu(t)$, we can compute the required derivative, $\dt L$ as,
\begin{equation}
   \dt L = - \int_0^T \lambda^T(t)\partial_{\theta}f_{NN}(u(t), y(t), t; \theta)dt\,.
\end{equation}

Now, we find the derivative of the Lagrangian w.r.t $\phi$, 
\begin{equation}
\begin{split}
   \dph L =& \partial_{u(t)} \mathcal{L}(u(t))\dph u(t) + \int_0^T \lm^T(t) \left(\dph d_t u(t) - \partial_{u(t)} f_{NN}(u(t), y(t), t; \theta)\dph u(t) \right. \\
   & \left. - \partial_{y(t)} f_{NN}(u(t), y(t), t;\theta) \dph y(t) \right)dt  + \int_0^T \mu^T(t) \left(\dph d_t y(t) \right. \\
   & \left. - \partial_{u(t - \tau_1)}g_{NN}(u(t - \tau_1), t-\tau_1; \phi)\dph u(t - \tau_1) + \partial_{u(t-\tau_2)}g_{NN}(u(t-\tau_2), t-\tau_2; \phi)\dph u(t - \tau_2) \right. \\
   & \left. - \partial_{\phi}g_{NN}(u(t - \tau_1), t-\tau_1; \phi) + \partial_{\phi}g_{NN}(u(t-\tau_2), t-\tau_2; \phi)\right)dt + \alpha^T \dph y(0) \\
   & - \alpha^T\int_{-\tau_2}^{-\tau_1} \partial_{\phi} g_{NN}(h(t), t; \phi)dt \,.
   \end{split}
   \label{eq: derivative lagrangian phi nDDE-Dist}
\end{equation}
using integration-by-parts and change of variables, we can write,
\begin{equation}
\begin{split}
   \int_0^T & \lm^T(t) \dph d_t u(t) dt = \lm^T(T) \dph u(T) - \lm^T(0)\dph u(0) - \int_0^T d_t\lm^T(t) \dph u(t) dt \; . \\
   \int_0^T & \mu^T(t) \dph d_t y(t) dt = \mu^T(T) \dph y(T) - \mu^T(0)\dph y(0) - \int_0^T d_t\mu^T(t) \dph y(t) dt \; .\\
   \int_0^T & \mu^T(t) \partial_{u(t - \tau_i)}g_{NN}(u(t - \tau_i), t-\tau_i; \phi) \dph u(t - \tau_i) dt  = \\ &  \qquad \qquad\qquad\qquad\qquad\qquad\qquad\qquad \int_{0}^{T}  \mu^T(t+\tau_i)\partial_{u(t)}g_{NN}(u(t), t; \phi) \dph u(t) dt  \\
   &\qquad \qquad\qquad\qquad\qquad\qquad\qquad\qquad + \int_{-\tau_i}^{0} \mu^T(t+\tau_i)\partial_{u(t)}g_{NN}(u(t), t; \phi) \dph u(t) dt \; .
\end{split}
\end{equation}
Substituting these in Eq.\;\ref{eq: derivative lagrangian phi nDDE-Dist}, and keeping in mind that the initial condition $h(t)$ is independent of $\phi$, we obtain,
\begin{equation}
\begin{split}
   d_{\phi} L =& \int_0^T\left(\sum_{i = 1}^M\partial_{u(t)} l(u(t))\delta(t - T_i) - d_t\lm^T(t) - \lm^T(t)\partial_{u(t)}f_{NN}(u(t), y(t), t; \theta) \right. \\
  &  - \mu^T(t + \tau_1) \partial_{u(t)}g_{NN}(u(t), t; \phi) +  \mu^T(t+\tau_2) \partial_{u(t)}g_{NN}(u(t), t; \phi)\bigg) \dph u(t) dt  \\
  & + \int_0^T \mu^T(t) \left( - \partial_{\phi}g_{NN}(u(t - \tau_1), t-\tau_1; \phi) + \partial_{\phi}g_{NN}(u(t-\tau_2), t-\tau_2; \phi)\right) dt \\
  &+ \int_0^T (-d_t\mu^T(t) - \lm^T(t)\partial_{y(t)}f_{NN}(u(t), y(t), t; \theta))\dph y(t) dt \\
   & - \mu^T(0) \dph y(0) + \alpha^T \dph y(0) - \alpha^T\int_{-\tau_2}^{-\tau_1} \partial_{\phi} g_{NN}(h(t), t; \phi)dt
   \end{split}
\end{equation}
As we already satisfy the adjoint equations (Eq.\;\ref{eq: adjoint equations nDDE-Dist}), and letting $\alpha^T=\mu^T(0)$, we arrive at the expression for $\dph L$,
\begin{equation}
\begin{split}
   \di L =& - \int_0^T \mu^T(t) \left( \partial_{\phi}g_{NN}(u(t - \tau_1), t-\tau_1; \phi) - \partial_{\phi}g_{NN}(u(t-\tau_2), t-\tau_2; \phi)\right) dt \\
   & - \mu^T(0)\int_{-\tau_2}^{-\tau_1} \partial_{\phi} g_{NN}(h(t), t; \phi)dt \,.
   \end{split}
\end{equation}


\section{Experimental Setup}
\label{SI: Experimental Setup}

\subsection{Architectures}
\label{SI: Architectures}
\begin{figure}
  \centering
  \subfloat[][Experiment-1]{\includegraphics[width=0.7\textwidth]{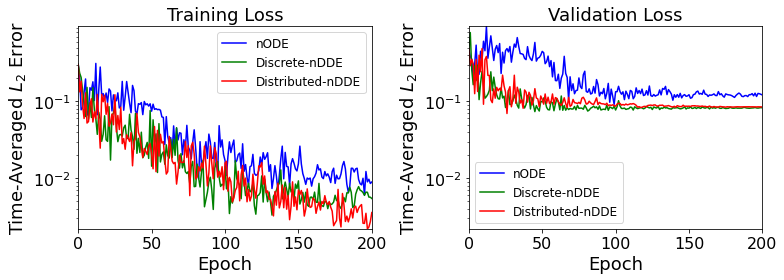}\label{fig: Exp_1_train_val_loss_all}} 
  \\
   \subfloat[][Experiment-2]{\includegraphics[width=0.7\textwidth]{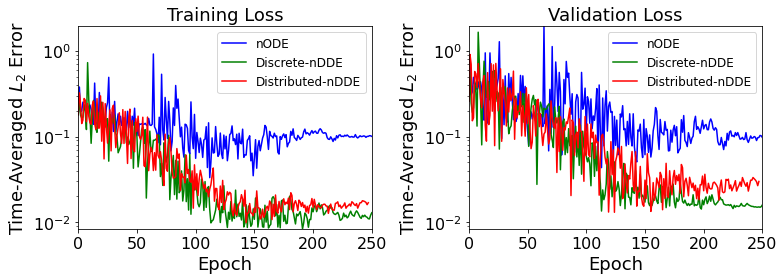}\label{fig: Exp_2_train_val_loss_all}}
  \\
   \subfloat[][Experiment-3a]{\includegraphics[width=0.7\textwidth]{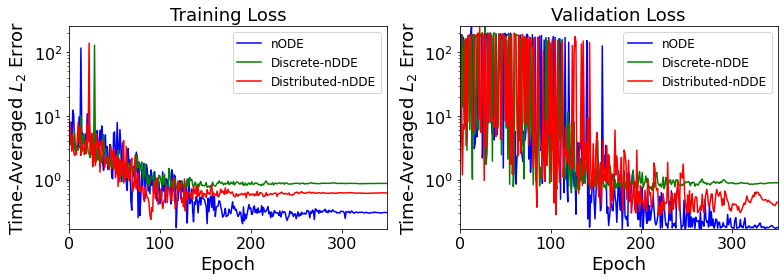}\label{fig: Exp_3_train_val_loss_all}}  
   \\
   \subfloat[][Experiment-3b]{\includegraphics[width=0.7\textwidth]{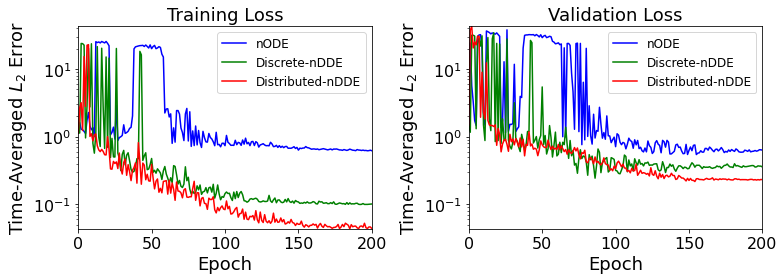}\label{fig: Exp_4_train_val_loss_all}}
  \caption{\small Variation with epochs of training (\textit{left column}), and validation (\textit{right column}) time-averaged $L_2$ loss for the three neural closure models, while training for each of the Experiments-1, 2, 3a, and 3b. These results accompany Figs.\;\ref{fig: Exp1_Comparison_Coeff_Plots},~\ref{fig: Exp2 Main results comparison between nODE, nDDE and nDistDDE},~\ref{fig: Exp3_Results_Combined},~\&~\ref{fig: Exp4 Main results comparison between nODE, nDDE and nDistDDE} in the main text, and the architectures detailed in Tables \ref{table: Exp1 and 2 architecture} \& \ref{table: Exp3 and 4 architecture} .}
  \label{fig: Loss variation for all experiments}
\end{figure}

In Tables\;\ref{table: Exp1 and 2 architecture} \& \ref{table: Exp3 and 4 architecture} we provide architectural details of the various neural closure models used in the main text. We also provide the variation of training and validation loss with training epochs corresponding to these architectures in Fig.\;\ref{fig: Loss variation for all experiments}. These results were picked among multiple repeats of training done with exactly the same hyperparameters (described next), with 3-5 repeats for discrete-nDDE in all the different experimental cases; 3-5 repeats for nODE and distributed-nDDE in experiments-1~\&~3b; and 10-15 repeats for nODE and distributed-nDDE in experiments-2~\&~3a (the same which were used for optimal delay length analysis).

\begin{table}
\resizebox{1\textwidth}{!}{%
\caption{\small
Architectures for different neural closure models used in Experiments-1 and 2. FC stands for fully-connected, Conv1D for convolutional-1D, and Conv1D-T for convolutional-1D transpose layers. The size of the convolutional layer filters is mentioned by the kernel size ($KS$; where the first dimension corresponds to the receptive field, and second to the number of channels), along with the number of strides ($S$).}
\label{table: Exp1 and 2 architecture}
\begin{tabular}{|p{0.15\textwidth}|p{0.2\textwidth}|p{0.05\textwidth}|p{0.15\textwidth}|p{0.15\textwidth}|p{0.3\textwidth}|p{0.05\textwidth}|p{0.15\textwidth}|p{0.15\textwidth}|}
\hline
\multirow{3}{*}{\textbf{Category}}      &       \multicolumn{4}{c|}{\textbf{Experiment - 1}}     &            \multicolumn{4}{c|}{\textbf{Experiment - 2}}       \\
\cline{2-5}     \cline{6-9}
&       \multirow{2}{*}{\textbf{Architecture}}      &       \multirow{2}{*}{\textbf{Act.}}      &       \multirow{2}{*}{\textbf{Delays}}        &       \multirow{2}{*}{\parbox{0.15\textwidth}{\textbf{Trainable Parameters}}}       &       \multirow{2}{*}{\textbf{Architecture}}      &       \multirow{2}{*}{\textbf{Act.}}      &       \multirow{2}{*}{\textbf{Delays}}        &       \multirow{2}{*}{\parbox{0.15\textwidth}{\textbf{Trainable Parameters}}}      \\
&       &       &       &       &       &       &       &       \\
\hline
\multirow{7}{*}{\parbox{0.15\textwidth}{nODE (No-Delays)}}      &       \multicolumn{2}{c|}{$f_{NN}$}        &       \multirow{7}{*}{None}       &       \multirow{7}{*}{158}     &       \multicolumn{2}{c|}{$f_{NN}$}        &       \multirow{7}{*}{None}       &       \multirow{7}{*}{424}        \\
\cline{2-3}     \cline{6-7}
&       Input layer with 3 neurons      &       none        &       &       &       Input of size $25 \times 1$      &       none        &       &       \\
\cline{2-3}     \cline{6-7}
&       5 FC hidden layer with 5 neurons        &       tanh        &       &       &       Conv1D layer with $KS = 3 \times 4$, $S = 1$        &       swish        &       &       \\
\cline{2-3}     \cline{6-7}
&       FC output layer with 3 neurons      &       linear      &       &       &       4 Conv1D layer with $KS = 3 \times 5$, $S = 1$      &       swish      &       &       \\
\cline{6-7}
&      &      &       &       &       Conv1D-T layer with $KS = 3 \times 3$, $S = 1$      &       swish      &       &       \\
\cline{6-7}
&      &      &       &       &       3 Conv1D-T layer with $KS = 3 \times 2$, $S = 1$      &       swish      &       &       \\
\cline{6-7}
&      &      &       &       &       Conv1D-T output layer with $KS = 3 \times 1$, $S = 1$      &       linear      &       &       \\
\hline
\multirow{6}{*}{\parbox{0.15\textwidth}{Discrete-nDDE}}      &       \multicolumn{2}{c|}{$f_{RNN}$}        &       \multirow{6}{*}{\parbox{0.125\textwidth}{$\tau_1 = 0.025$, $\tau_2 = 0.05$, ..., $\tau_6 = 0.15$}}       &       \multirow{6}{*}{63}     &       \multicolumn{2}{c|}{$f_{RNN}$}        &       \multirow{6}{*}{\parbox{0.125\textwidth}{$\tau_1 = 0.025$, $\tau_2 = 0.05$, ..., $\tau_6 = 0.15$}}       &       \multirow{6}{*}{110}        \\
\cline{2-3}     \cline{6-7}
&       Input layer with 3 neurons      &       none        &       &       &       Input of size $25\times1$      &       none        &       &       \\
\cline{2-3}     \cline{6-7}
&       Simple RNN layer with 5 neurons        &       tanh        &       &       &       Simple RNN Conv1D layer with $KS = 3 \times 3$, $S = 1$        &       swish        &       &       \\
\cline{2-3}     \cline{6-7}
&       FC output layer with 3 neurons      &       linear      &       &       &       Conv1D layer with $KS = 3 \times 2$, $S = 1$      &       swish      &       &       \\
\cline{6-7}
&       &      &       &       &       Conv1D-T layer with $KS = 3 \times 2$, $S = 1$      &       swish      &       &       \\
\cline{6-7}
&       &      &       &       &       Conv1D-T output layer with $KS = 3 \times 1$, $S = 1$      &       linear      &       &       \\
\hline
\multirow{12}{*}{\parbox{0.15\textwidth}{Distributed-nDDE}}      &       \multicolumn{2}{c|}{$f_{NN}$}        &       \multirow{12}{*}{\parbox{0.125\textwidth}{$\tau_1 = 0.0$, $\tau_2 = 0.075$}}       &       \multirow{12}{*}{110}     &       \multicolumn{2}{c|}{$f_{NN}$}        &       \multirow{12}{*}{\parbox{0.125\textwidth}{$\tau_1 = 0.0$, $\tau_2 = 0.075$}}       &       \multirow{12}{*}{361}        \\
\cline{2-3}     \cline{6-7}
&       Input layer with 5 neurons      &       none        &       &       &       Input of size $25 \times 2$      &       none        &       &       \\
\cline{2-3}     \cline{6-7}
&       2 FC hidden layer with 5 neurons        &       tanh        &       &       &        Conv1D layer with $KS = 3 \times 4$, $S = 1$        &       swish        &       &       \\
\cline{2-3}     \cline{6-7}
&       FC output layer with 3 neurons      &       linear      &       &       &       2 Conv1D layer with $KS = 3 \times 5$, $S = 1$      &       swish      &       &       \\
\cline{6-7}
&       &      &       &       &       Conv1D-T layer with $KS = 3 \times 3$, $S = 1$      &       swish      &       &       \\
\cline{6-7}
&       &      &       &       &       Conv1D-T layer with $KS = 3 \times 2$, $S = 1$      &       swish      &       &       \\
\cline{6-7}
&       &      &       &       &       Conv1D-T output layer with $KS = 3 \times 1$, $S = 1$      &       linear      &       &       \\
\cline{2-3}     \cline{6-7}
&       \multicolumn{2}{c|}{$g_{NN}$}      &       &       &       \multicolumn{2}{c|}{$g_{NN}$}      &       &       \\
\cline{2-3}     \cline{6-7}
&       Input layer with 3 neurons      &       none        &       &       &       Input of size $25\times1$      &       none        &       &       \\
\cline{2-3}     \cline{6-7}
&       2 FC hidden layer with 3 neurons        &       tanh        &       &       &       Conv1D layer with $KS = 3 \times 2$, $S = 1$        &       swish        &       &       \\
\cline{2-3}     \cline{6-7}
&       FC output layer with 2 neurons      &       linear      &       &       &       Conv1D layer with $KS = 3 \times 3$, $S = 1$      &       swish      &       &       \\
\cline{6-7}
&       &      &       &       &       Conv1D-T layer with $KS = 3 \times 3$, $S = 1$      &       swish      &       &       \\
\cline{6-7}
&       &      &       &       &       Conv1D-T output layer with $KS = 3 \times 1$, $S = 1$      &       linear      &       &       \\
\hline
\end{tabular}}
\end{table}


\begin{table}
\resizebox{1\textwidth}{!}{%
\caption{\small
Architectures for different neural closure models used in Experiments-3a and 3b. FC stands for fully-connected, and Conv1D for convolutional-1D layers. The size of the convolutional layer filters is mentioned by the kernel size ($KS$; where the first dimension corresponds to the receptive field, and second to the number of channels), along with the number of strides ($S$). While \textit{AddExtraChannels} and \textit{BioConstrainLayer} are custom layers described in the main text (Secs.\;\ref{sec:biogeochemical 0-D models}~\&~\ref{sec:biogeochemical 1-D models}).}
\label{table: Exp3 and 4 architecture}
\begin{tabular}{|p{0.15\textwidth}|p{0.2\textwidth}|p{0.05\textwidth}|p{0.15\textwidth}|p{0.15\textwidth}|p{0.3\textwidth}|p{0.05\textwidth}|p{0.15\textwidth}|p{0.15\textwidth}|}
\hline
\multirow{3}{*}{\textbf{Category}}      &       \multicolumn{4}{c|}{\textbf{Experiment - 3a}}     &       \multicolumn{4}{c|}{\textbf{Experiment - 3b}}       \\
\cline{2-5}     \cline{6-9}
&       \multirow{2}{*}{\textbf{Architecture}}      &       \multirow{2}{*}{\textbf{Act.}}      &       \multirow{2}{*}{\textbf{Delays}}        &       \multirow{2}{*}{\parbox{0.15\textwidth}{\textbf{Trainable Parameters}}}      &       \multirow{2}{*}{\textbf{Architecture}}      &       \multirow{2}{*}{\textbf{Act.}}      &       \multirow{2}{*}{\textbf{Delays}}        &       \multirow{2}{*}{\parbox{0.15\textwidth}{\textbf{Trainable Parameters}}}      \\
&       &       &       &       &       &       &       &       \\
\hline
\multirow{5}{*}{\parbox{0.15\textwidth}{nODE (No-Delays)}}      &       \multicolumn{2}{c|}{$f_{NN}$}        &       \multirow{5}{*}{None}       &       \multirow{5}{*}{317}     &       \multicolumn{2}{c|}{$f_{NN}$}        &       \multirow{5}{*}{None}       &       \multirow{5}{*}{987}        \\
\cline{2-3}     \cline{6-7}
&       Input layer with 3 neurons      &       none        &       &       &       Input of size $20 \times 3$      &       none        &       &       \\
\cline{2-3}     \cline{6-7}
&       6 FC hidden layer with 7 neurons        &       tanh        &       &       &       AddExtraChannels\{z, I(z, t)\}        &       none        &       &       \\
\cline{2-3}     \cline{6-7}
&       FC hidden layer with 1 neuron      &       linear      &       &       &        Conv1D layers with $KS= 1\times5$, $S= 1$;  $KS= 1\times7$, $S= 1$;  $KS= 1\times9$, $S= 1$;  $KS= 1\times11$, $S= 1$;  $KS= 1\times13$, $S= 1$;  $KS= 1\times13$, $S= 1$;  $KS= 1\times11$, $S= 1$;  $KS= 1\times9$, $S= 1$;  $KS= 1\times7$, $S= 1$;  $KS= 1\times5$, $S= 1$;  $KS= 1\times3$, $S= 1$        &       swish      &       &       \\
\cline{2-3}     \cline{6-7}
&       BioConstrainLayer with output of size 3      &       linear      &       &       &         Conv1D layer with $KS= 1\times1$, $S= 1$      &       linear      &       &       \\
\cline{6-7}
&       &      &       &       &       BioConstrainLayer with output of size $20 \times 3$      &       linear      &       &       \\
\hline
\multirow{6}{*}{\parbox{0.15\textwidth}{Discrete-nDDE}}      &       \multicolumn{2}{c|}{$f_{RNN}$}        &       \multirow{6}{*}{\parbox{0.125\textwidth}{$\tau_1 = 0.75$, $\tau_2 = 1.5$, ..., $\tau_6 = 4.5$}}       &       \multirow{6}{*}{142}     &       \multicolumn{2}{c|}{$f_{RNN}$}        &       \multirow{6}{*}{\parbox{0.125\textwidth}{$\tau_1 = 0.5$, $\tau_2 = 1.0$, $\tau_3 = 1.5$, $\tau_4 = 2.0$}}       &       \multirow{6}{*}{426}        \\
\cline{2-3}     \cline{6-7}
&       Input layer with 3 neurons      &       none        &       &       &       Input of size $20\times 3$      &       none        &       &       \\
\cline{2-3}     \cline{6-7}
&       Simple RNN layer with 7 neurons        &       tanh        &       &       &        Simple  RNN  Conv1D  layer  with $KS= 1\times5$, $S= 1$       &       swish
&       &       \\
\cline{2-3}     \cline{6-7}
&       FC hidden layer with 7 neurons      &       tanh      &       &       &       AddExtraChannels\{z, I(z, t)\}      &       none      &       &       \\
\cline{2-3}     \cline{6-7}
&   FC hidden layer with 1 neuron    &  linear    &       &       &       Conv1D  layers  with $KS= 1\times7$, $S= 1$; $KS= 1\times9$, $S= 1$; $KS= 1\times9$, $S= 1$; $KS= 1\times7$, $S= 1$; $KS= 1\times5$, $S= 1$; $KS= 1\times3$, $S= 1$      &       swish      &       &       \\
\cline{2-3}     \cline{6-7}
&   BioConstrainLayer with output of size 3    &  linear    &       &       &       Conv1D  layer  with $KS= 1\times1$, $S= 1$      &       linear      &       &       \\
\cline{6-7}
&       &      &       &       &       BioConstrainLayer with output of size $20 \times 3$      &       linear      &       &       \\
\hline
\multirow{9}{*}{\parbox{0.15\textwidth}{Distributed-nDDE}}      &       \multicolumn{2}{c|}{$f_{NN}$}        &       \multirow{9}{*}{\parbox{0.125\textwidth}{$\tau_1 = 0.0$, $\tau_2 = 2.5$}}       &       \multirow{9}{*}{195}       &       \multicolumn{2}{c|}{$f_{NN}$}        &       \multirow{9}{*}{\parbox{0.125\textwidth}{$\tau_1 = 0.0$, $\tau_2 = 2.0$}}       &       \multirow{9}{*}{477}        \\
\cline{2-3}     \cline{6-7}
&       Input layer with 11 neurons      &       none        &       &       &       Input of size $20\times 5$      &       none        &       &       \\
\cline{2-3}     \cline{6-7}
&       2 FC hidden layer with 7 neurons        &       tanh        &       &       &       AddExtraChannels\{z, I(z, t)\}        &       none        &       &       \\
\cline{2-3}     \cline{6-7}
&       FC hidden layer with 1 neurons      &       linear      &       &       &       Conv1D  layers  with $KS= 1\times7$, $S= 1$; $KS= 1\times9$, $S= 1$; $KS= 1\times9$, $S= 1$; $KS= 1\times7$, $S= 1$; $KS= 1\times5$, $S= 1$; $KS= 1\times3$, $S= 1$      &       swish      &       &       \\
\cline{2-3}     \cline{6-7}
&       BioConstrainLayer with output of size 3      &       linear      &       &       &       Conv1D  layer  with $KS= 1\times1$, $S= 1$      &       linear      &       &       \\
\cline{6-7}
&       &       &       &       &       BioConstrainLayer with output of size $20\times 3$      &       linear      &       &       \\
\cline{2-3}     \cline{6-7}
&       \multicolumn{2}{c|}{$g_{NN}$}      &       &       &       \multicolumn{2}{c|}{$g_{NN}$}      &       &       \\
\cline{2-3}     \cline{6-7}
&       Input layer with 3 neurons      &       none        &       &       &       Input of size $20\times 3$      &       none        &       &       \\
\cline{2-3}     \cline{6-7}
&       2 FC hidden layer with 5 neurons        &       tanh        &       &       &       Conv1D  layers  with $KS= 1\times3$, $S= 1$; $KS= 1\times5$, $S= 1$; $KS= 1\times7$, $S= 1$; $KS= 1\times5$, $S= 1$       &       swish        &       &       \\
\cline{2-3}     \cline{6-7}
&       FC output layer with 4 neurons      &       linear      &       &       &       Conv1D  output layer  with $KS= 1\times2$, $S= 1$      &       linear      &       &       \\
\hline
\end{tabular}}
\end{table}


\subsection{Hyperparameters}
\label{SI: hyperparameters}
The values of the various training hyperparameters used in the experiments are listed next. In all the experiments, the number of iterations per epoch are calculated by dividing the number of time-steps in the training period by batch-size multiplied the length of short time-sequences, adding 1, and rounding up to the next integer.

\textbf{Experiments-1:} For training, we randomly select short time-sequences spanning only 6 time-steps and extract data at every other time-step to form batches of size 2; 18 iterations per epoch; exponentially decaying learning rate (LR) schedule (initial LR of 0.075, decay rate of 0.97, and 18 decay steps); \textit{RMSprop} optimizer; and end training at 200 epochs.

\textbf{Experiments-2:} We use a batch size of 8 created by randomly selecting short time-sequences spanning 6 time-steps and extracting data at every other time-step; 4 iterations per epoch; exponentially decaying learning rate (LR) schedule with initial LR of 0.075, decay rate of 0.97, and 4 decay steps; \textit{RMSprop} optimizer; and end training at 250 epochs. 

\textbf{Experiments-3a:} We use a batch size of 4 created by randomly selecting short time-sequences spanning 6 time-steps and extracting data at every other time-step; 26 iterations per epoch; exponentially decaying learning rate (LR) schedule with initial LR of 0.05, decay rate of 0.97, and 26 decay steps; \textit{RMSprop} optimizer; and end training at 350 epochs.

\textbf{Experiments-3b:} We use a batch size of 8 (4 for only distributed-nDDE) created by randomly selecting short time-sequences spanning 6 time-steps and extracting data at every other time-step; 8 (14 for only distributed-nDDE) iterations per epoch; exponentially decaying learning rate (LR) schedule with initial LR of 0.05, decay rate of 0.97, and 8 (14 for only distributed-nDDE) decay steps; \textit{RMSprop} optimizer; and end training at 200 epochs.


\subsection{Sensitivity to Network Size and Training Period Length}
\label{SI: Sensitivity to Network Size and Training Period Length}
We performed various hyperparameter studies for all the different experimental cases presented in the main paper. However, here we only show the effect of network size and training period length on the performance of experiments-1 with distributed-nDDE closure. 

\begin{figure}[h!]
  \centering
  \subfloat[][Effect on performance for change in training period length]{\includegraphics[width=0.6\textwidth]{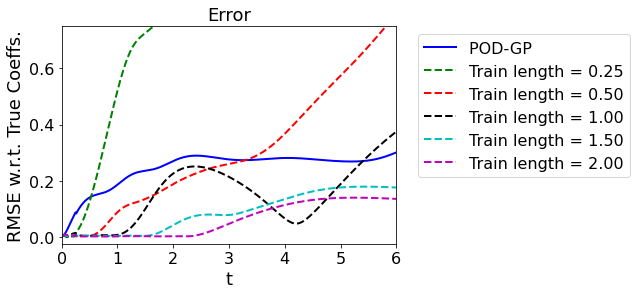}\label{fig: Exp1_RMSE_with_time_train_time}} \\
  \subfloat[][Change in time-average $L_2$ loss with change in the network size]{\includegraphics[width=0.8\textwidth]{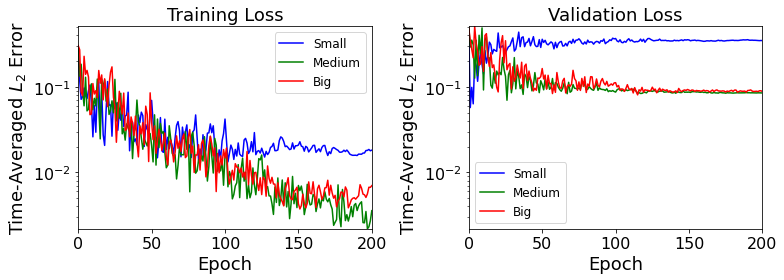}\label{fig: Exp_1_train_val_loss_all_network_size}}
  \caption{
  Experiments-1 sensitivity to network size and training period length. {(a):} Evolution of root-mean-squared-error (RMSE) of coefficients for distributed-nDDEs trained with different training period length, and with same architectures and other hyperparameter values. These results correspond to the distributed-nDDE architecture detailed in Table\;\ref{table: Exp1 and 2 architecture}. {(b):} Variation with epochs of training (\textit{left}), and validation (\textit{right}) time-averaged $L_2$ loss for the three different sized distributed-nDDE architectures detailed in Table~\ref{table: Exp1 different architecture sizes}.
  }
  \label{fig:Exp1 train time and network size}
\end{figure}

First, we varied the length of the training period while keeping the architecture and other hyperparameters the same (as mentioned above, Secs.\;\ref{SI: Architectures}~\&~\ref{SI: hyperparameters}). We chose 5 different training lengths (all starting from $t=0$), with the longest encompassing one time-period for the coefficient corresponding to mode 3. In Fig.\;\ref{fig: Exp1_RMSE_with_time_train_time}, we provide the root-mean-square-error (RMSE) as it evolves with time, spanning the training, validation, and prediction periods. We can notice, that in each case, the network is able to exactly match the true coefficients upto the end of training period. While the long-term performance drastically improves on providing more-and-more training data by increasing the training period length.

Second, we only vary the depth of the network, while keeping all other training details exactly the same (as mentioned above, Sec.\;\ref{SI: hyperparameters}). We chose three different network sizes by changing the number of hidden layers, with architectural details presented in Table\;\ref{table: Exp1 different architecture sizes}. Often, smaller networks are underparameterized limiting their expressivity, while overly large networks might become overparameterized limiting their generalizability for predictions. In Fig.\;\ref{fig: Exp_1_train_val_loss_all_network_size}, we provide variation of training and validation losses with training epochs, and can notice that the small network struggles to close the system.

\begin{table}
\resizebox{0.65\textwidth}{!}{%
\caption{\small
Architectures of different sizes for distributed-nDDE used in hyperparameter sensitivity study for Experiments-1.}
\label{table: Exp1 different architecture sizes}
\begin{tabular}{|m{0.15\textwidth}|m{0.35\textwidth}|m{0.05\textwidth}|m{0.15\textwidth}|m{0.15\textwidth}|}
\hline
\textbf{Category}      &              \textbf{Architecture}      &      \textbf{Act.}      &       \textbf{Delays}        &       \textbf{Trainable Parameters}       \\
\hline
\multirow{6}{*}{\parbox{0.15\textwidth}{Small}}      &       \multicolumn{2}{c|}{$f_{NN}$}        &       \multirow{6}{*}{\parbox{0.1\textwidth}{$\tau_1 = 0.0$, $\tau_2 = 0.075$}}       &       \multirow{6}{*}{38}             \\
\cline{2-3}     
&       Input layer with 5 neurons      &       none        &       &              \\
\cline{2-3}    
&       FC output layer with 3 neurons      &       linear      &       &              \\
\cline{2-3}     
&       \multicolumn{2}{c|}{$g_{NN}$}       &       &              \\
\cline{2-3}     
&       Input layer with 3 neurons      &       none        &       &              \\
\cline{2-3}     
&       FC hidden layer with 3 neurons        &       tanh        &       &              \\
\cline{2-3}     
&       FC output layer with 2 neurons      &       linear      &       &              \\
\hline
\multirow{8}{*}{\parbox{0.15\textwidth}{Medium}}      &       \multicolumn{2}{c|}{$f_{NN}$}        &       \multirow{8}{*}{\parbox{0.1\textwidth}{$\tau_1 = 0.0$, $\tau_2 = 0.075$}}       &       \multirow{8}{*}{110}             \\
\cline{2-3}     
&       Input layer with 5 neurons      &       none        &       &              \\
\cline{2-3}     
&       2 FC hidden layer with 5 neurons        &       tanh        &       &              \\
\cline{2-3}    
&       FC output layer with 3 neurons      &       linear      &       &              \\
\cline{2-3}     
&       \multicolumn{2}{c|}{$g_{NN}$}       &       &              \\
\cline{2-3}     
&       Input layer with 3 neurons      &       none        &       &              \\
\cline{2-3}     
&       2 FC hidden layer with 3 neurons        &       tanh        &       &              \\
\cline{2-3}     
&       FC output layer with 2 neurons      &       linear      &       &              \\
\hline
\multirow{8}{*}{\parbox{0.15\textwidth}{Big}}      &       \multicolumn{2}{c|}{$f_{NN}$}        &       \multirow{8}{*}{\parbox{0.1\textwidth}{$\tau_1 = 0.0$, $\tau_2 = 0.075$}}       &       \multirow{8}{*}{152}             \\
\cline{2-3}     
&       Input layer with 5 neurons      &       none        &       &              \\
\cline{2-3}     
&       3 FC hidden layer with 5 neurons        &       tanh        &       &              \\
\cline{2-3}    
&       FC output layer with 3 neurons      &       linear      &       &              \\
\cline{2-3}     
&       \multicolumn{2}{c|}{$g_{NN}$}       &       &              \\
\cline{2-3}     
&       Input layer with 3 neurons      &       none        &       &              \\
\cline{2-3}     
&       3 FC hidden layer with 3 neurons        &       tanh        &       &              \\
\cline{2-3}     
&       FC output layer with 2 neurons      &       linear      &       &              \\
\hline
\end{tabular}
}
\end{table}

\bibliographystyle{unsrtnat}
\bibliography{mseas,references}

\makeatletter\@input{suppxx.tex}\makeatother